\documentclass[letterpaper]{article} % DO NOT CHANGE THIS
\usepackage{aaai2026}  % DO NOT CHANGE THIS
\usepackage{times}  % DO NOT CHANGE THIS
\usepackage{helvet}  % DO NOT CHANGE THIS
\usepackage{courier}  % DO NOT CHANGE THIS
\usepackage[hyphens]{url}  % DO NOT CHANGE THIS
\usepackage{graphicx} % DO NOT CHANGE THIS
\usepackage{natbib}  % DO NOT CHANGE THIS AND DO NOT ADD ANY OPTIONS TO IT
\usepackage{caption} % DO NOT CHANGE THIS AND DO NOT ADD ANY OPTIONS TO IT
\usepackage{algorithm}
\usepackage{algorithmic}
\usepackage[utf8]{inputenc}

\usepackage{amsmath}

\usepackage{newfloat}
\usepackage{listings}

\usepackage[utf8]{inputenc} % allow utf-8 input
\usepackage[T1]{fontenc}    % use 8-bit T1 fonts
\usepackage{url}            % simple URL typesetting
\usepackage{booktabs}       % professional-quality tables
\usepackage{amsfonts}       % blackboard math symbols
\usepackage{nicefrac}       % compact symbols for 1/2, etc.
\usepackage{microtype}      % microtypography
\usepackage{xcolor}         % colors

\usepackage{algorithm}
\usepackage{algorithmic}
\usepackage{multirow}
\usepackage{bm}
\usepackage{amsmath,amssymb,amsfonts}
\usepackage{epsfig}
\usepackage{colortbl}

\usepackage{amsthm}
\usepackage{tablefootnote}
\usepackage{subcaption}
\usepackage{tcolorbox}
\theoremstyle{plain}
\newtheorem{theorem}{Theorem}

\theoremstyle{definition}

\theoremstyle{remark}

\usepackage{newfloat}
\usepackage{listings}
\DeclareCaptionStyle{ruled}{labelfont=normalfont,labelsep=colon,strut=off} % DO NOT CHANGE THIS
\floatstyle{ruled}
\newfloat{listing}{tb}{lst}{}
\floatname{listing}{Listing}
\title{Put the Space of LoRA Initialization to the Extreme \\ to Preserve Pre-trained Knowledge}
\author{
    Pengwei Tang\textsuperscript{\rm 1,2,3},
    Xiaolin Hu\textsuperscript{\rm 4} \thanks{Corresponding Author.},
    Yong Liu\textsuperscript{\rm 1,2,3}, \\
    Lizhong Ding\textsuperscript{\rm 5}, 
    Dongjie Zhang\textsuperscript{\rm 6},
    Xing Wu\textsuperscript{\rm 6,7},
    Debing Zhang \textsuperscript{\rm 6}
}

\affiliations{
    \textsuperscript{\rm 1}Gaoling School of Artificial Intelligence, Renmin University of China, Beijing, China, \\
    \textsuperscript{\rm 2} Beijing Key Laboratory of Research on Large Models and Intelligent Governance, \\
    \textsuperscript{\rm 3}Engineering Research Center of Next-Generation Intelligent Search and Recommendation, MOE,\\
    \textsuperscript{\rm 4}Key Laboratory of Multimedia Trusted Perception and Efficient Computing, Ministry of Education of China, Xiamen University,\\
    \textsuperscript{\rm 5}Beijing Institute of Technology, \\
    \textsuperscript{\rm 6}Xiaohongshu Inc., \\
    \textsuperscript{\rm 7}University of Chinese Academy of Sciences, Beijing, China \\
    tangpwei@ruc.edu.cn, xiaolinhu@xmu.edu.cn%, \{zhangdongjie, yuchen5\}@xiaohongshu.com, wuxing@iie.ac.cn
}

\usepackage{bibentry}
\begin{document}
\maketitle
\begin{abstract}
Low-Rank Adaptation (LoRA) is the leading parameter-efficient fine-tuning method for Large Language Models (LLMs), but it still suffers from catastrophic forgetting. Recent work has shown that specialized LoRA initialization can alleviate catastrophic forgetting. There are currently two approaches to LoRA initialization aimed at preventing knowledge forgetting during fine-tuning: (1) making residual weights close to pre-trained weights, and (2) ensuring the space of LoRA initialization is orthogonal to pre-trained knowledge. The former is what current methods strive to achieve, while the importance of the latter is not sufficiently recognized. We find that the space of LoRA initialization is the key to preserving pre-trained knowledge rather than the residual weights. Existing methods like MiLoRA propose making the LoRA initialization space orthogonal to pre-trained weights. However, MiLoRA utilizes the null space of pre-trained weights. Compared to pre-trained weights, the input activations of pre-trained knowledge take into account the parameters of all previous layers as well as the input data, while pre-trained weights only contain information from the current layer. Moreover, we find that the effective ranks of input activations are much smaller than those of pre-trained weights. Thus, the null space of activations is more accurate and contains less pre-trained knowledge information compared to that of weights. Based on these, we introduce LoRA-Null, our proposed method that initializes LoRA in the null space of activations. Experimental results show that LoRA-Null effectively preserves the pre-trained world knowledge of LLMs while achieving good fine-tuning performance, as evidenced by extensive experiments. %Code is available at {https://github.com/HungerPWAY/LoRA-Null}.
\end{abstract}

\begin{links}
    \link{Code}{https://github.com/HungerPWAY/LoRA-Null}
    %\link{Datasets}{https://aaai.org/example/datasets}
    %\link{Extended version}{https://arxiv.org/abs/2503.02659}
\end{links}

\section{Introduction}

Low-rank adaptation (LoRA), which has already become the leading parameter-efficient fine-tuning (PEFT) \cite{prompt_tuning,li-liang-2021-prefix,adapter,hu2022lora,tang2025adept,ijcai2025p760} method for Large Language Models (LLMs), assumes weight changes during fine-tuning have a low-rank structure \cite{hu2022lora}. 
To fine-tune pre-trained weight $\mathbf{W}_0 \in \mathbb{R}^{n \times m}$, LoRA uses two low-rank matrices $\mathbf{A} \in \mathbb{R}^{r \times m}$ and $\mathbf{B} \in \mathbb{R}^{n \times r}$ with $r \ll \min(m,n)$, to represent the corresponding weight change, \textit{i.e.}, $\mathbf{\Delta W} = \mathbf{B} \mathbf{A}$. The vanilla LoRA uses random Gaussian to initialize the down-projection matrix $\mathbf{A}$ and zero to initialize the up-projection matrix $\mathbf{B}$, which makes $\mathbf{B} \mathbf{A} = \mathbf{0}$ at the beginning of fine-tuning. LoRA reduces trainable parameters by tuning only low-rank matrices. 

Recent studies \cite{loralearnless} show that LoRA suffers less from this issue compared to full fine-tuning, but it still exhibits significant catastrophic forgetting \cite{loralearnless,loramoe,llamapro}. For LoRA fine-tuning, catastrophic forgetting can be mitigated through appropriate initialization techniques, such as CorDA \cite{yang2024corda} and MiLoRA \cite{milora}. These LoRA initialization methods use not zero and not random Gaussian initialization to initialize $\mathbf{A}$ and $\mathbf{B}$. Unlike vanilla LoRA, they require the base weights to be adjusted such that 
the initial merged weights $\mathbf{W}_0 + \mathbf{B} \mathbf{A}$ equal the original weights. 
The adjusted base weights, referred to as the \emph{residual weights}, are denoted by
$\mathbf{W}_0' = \mathbf{W}_0 - \mathbf{B} \mathbf{A}$. 

\begin{figure*}[t]
    \centering

    \begin{subfigure}{0.33\textwidth}
        \centering
        \includegraphics[width=\linewidth]{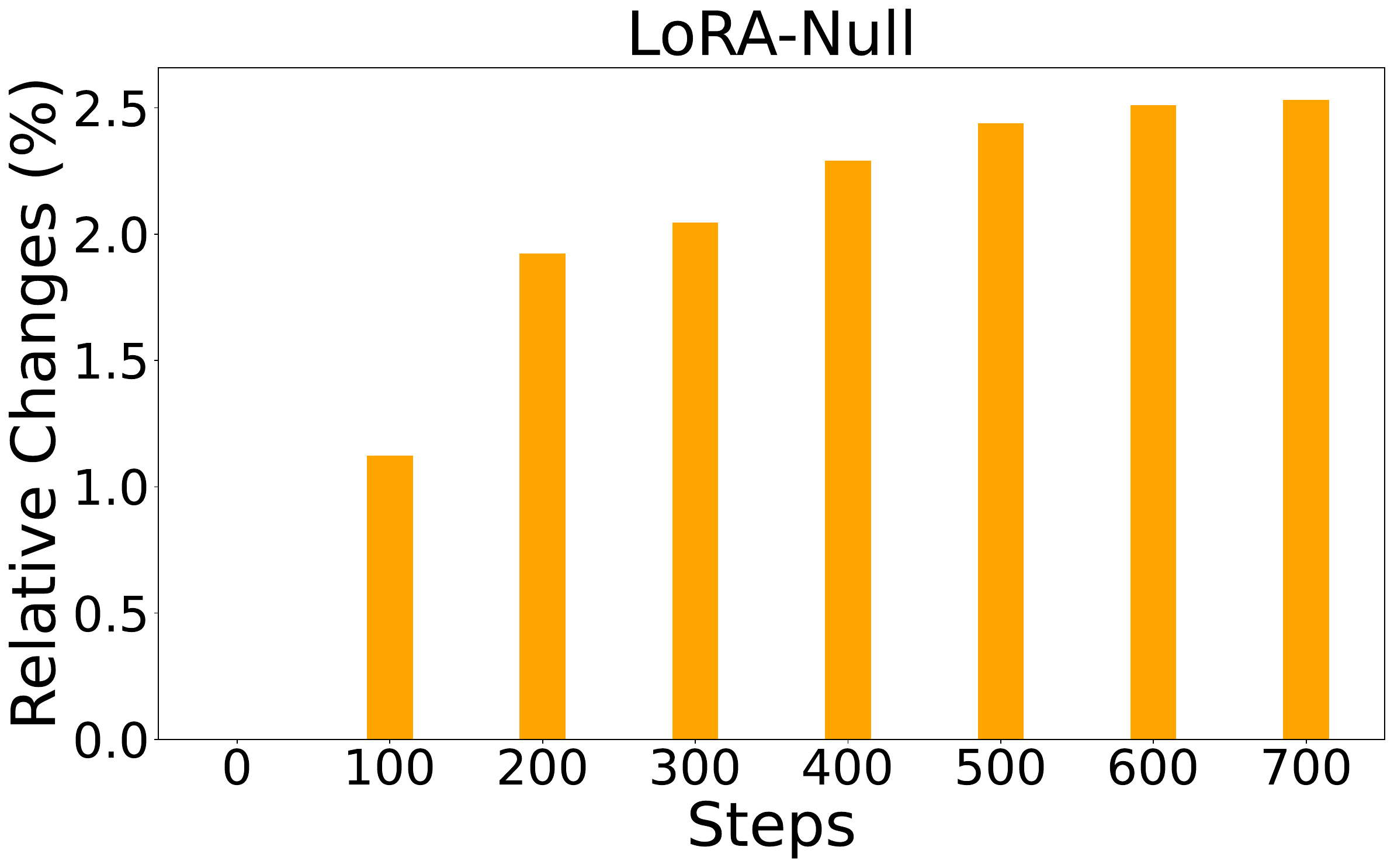}
        %\caption{LoRA-Null}
        \label{fig:llama3.2_0_k_lora_null}
    \end{subfigure}%
    \hfill
    \begin{subfigure}{0.33\textwidth}
        \centering
        \includegraphics[width=\linewidth]{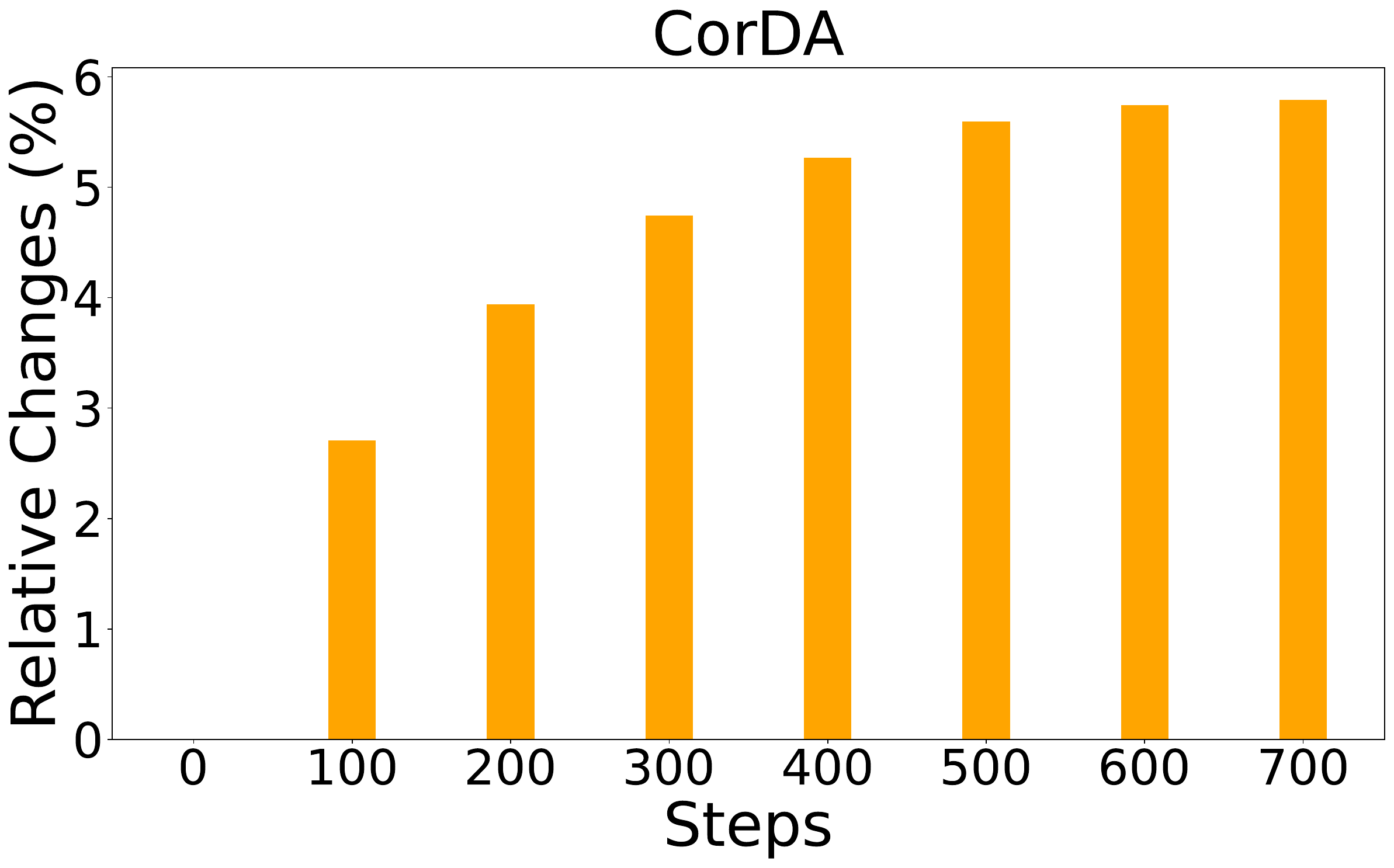}
        %\caption{CorDA}
        \label{fig:llama3.2_0_k_corda}
    \end{subfigure}%
    \hfill
    \begin{subfigure}{0.33\textwidth}
        \centering
        \includegraphics[width=\linewidth]{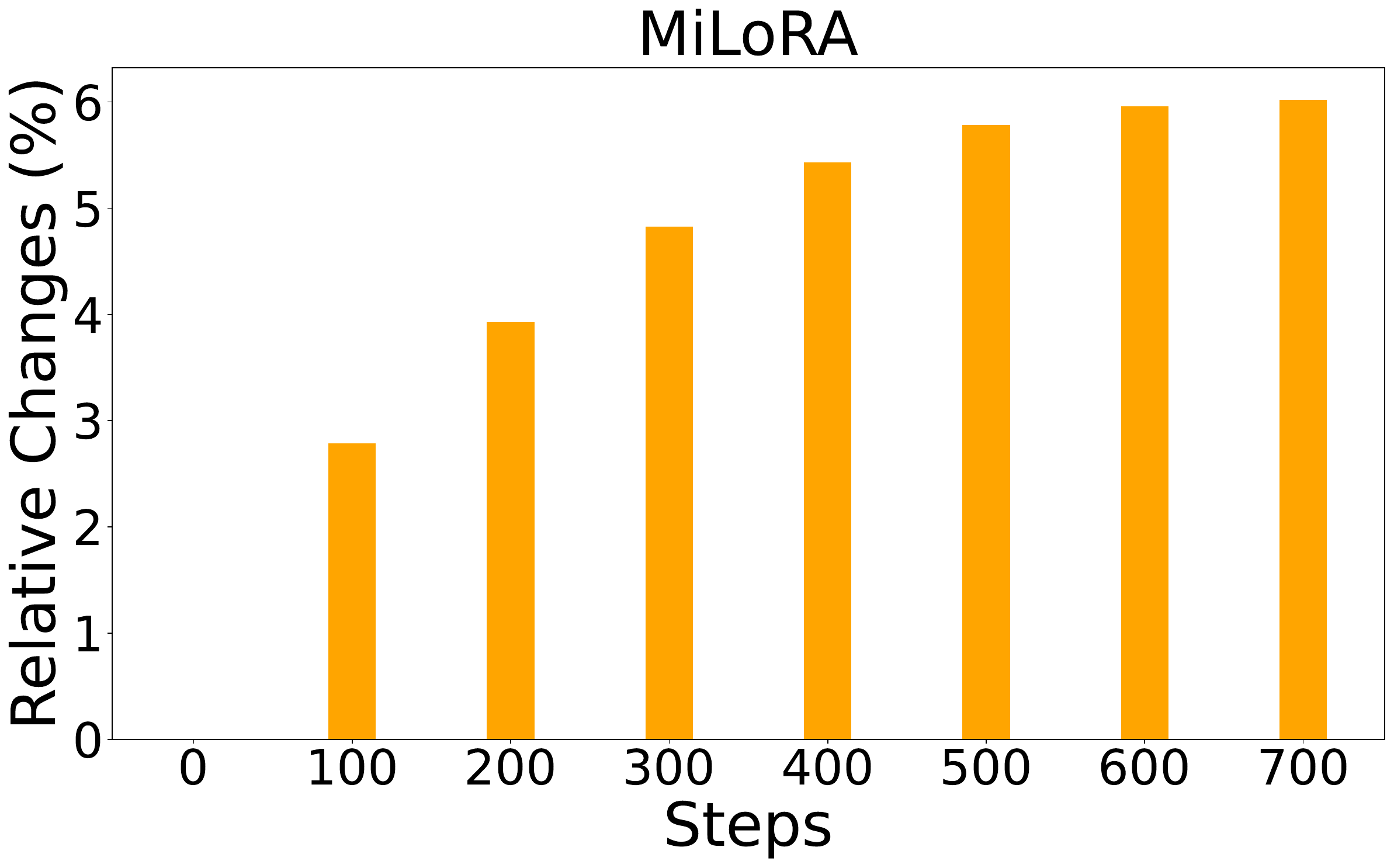}
        %\caption{MiLoRA}
        \label{fig:llama3.2_0_k_milora}
    \end{subfigure}

    \caption{The relative changes of LoRA-Null, CorDA and MiLoRA. It is caculated by $\|\mathbf{A}^*-\mathbf{A}_0\|_F/ \|\mathbf{A}_0\|_F$, where $\mathbf{A}^*$ is the tuned parameters. We calculate the relative changes on the $\mathbf{A}$ matrix of the key projection of layer 0 on LLaMA-3.2-3B.}
    \label{fig:llama3.2_relative}
\end{figure*}

\begin{table}[t]
\centering
\small
\setlength{\tabcolsep}{1mm}
\begin{tabular}{l|c|c}
\toprule
\textbf{Method} & 
\shortstack[c] { Making  $\mathbf{W}'_0 $ \\ close to $\mathbf{W}_0$} &
\shortstack[c]{$\mathbf{B}\mathbf{A}$ orthogonal \\ to Pretraind Knowlegde} \\
\midrule
LoRA       & \checkmark & \text{\sffamily X} \\
MiLoRA     & \checkmark & \checkmark \\
CorDA      & \checkmark & \text{\sffamily X} \\
LoRA-Null  & \text{\sffamily X} & \checkmark \\
\bottomrule
\end{tabular}
\caption{Comparison of methods in residual weights and LoRA initialization space.}
\label{tab:methods_comparsions}
\end{table}

MiLoRA \cite{milora} updates the minor singular components of the pre-trained weights, preserving pre-trained knowledge by freezing the principal components and using the less-optimized subspace for learning. CorDA \cite{yang2024corda} conducts Singular Value Decomposition (SVD) \cite{strang2022introduction} on pre-trained weights guided by the covariance matrix of pre-trained knowledge activations, integrating the context into the principal components to keep knowledge.

Based on the observations from MiLoRA and CorDA, we can summarize that there are currently two approaches to LoRA initialization aimed at preventing knowledge forgetting during fine-tuning: (1) making residual weights close to pre-trained weights, which are adopted by both CorDA and MiLoRA; (2) ensuring the space of LoRA initialization is orthogonal to pre-trained knowledge, which are used by MiLoRA. In Table \ref{tab:methods_comparsions}, we show that comparison of methods in residual weights and the space of LoRA initializations. Because it is believed that LoRA adapters $\mathbf{B}\mathbf{A}$ will change during fine-tuning, keeping the frozen residual weights close to the pre-trained weights seems like the key factor that enables the preservation of pre-trained knowledge. However, for CorDA and MiLoRA, the LoRA initialization is the component of the pre-trained weights \cite{yang2024corda,milora}. Just as proper fine-tuning keeps weights close to pre-trained weights, fine-tuned LoRA adapters remain close to their initializations. In Figure \ref{fig:llama3.2_relative}, we show that the relative changes of LoRA adapters are small. This suggests that, for knowledge preservation, focusing the primary objective on residual weights may not be that important. \textbf{A supporting example is that vanilla LoRA, despite freezing the full pre-trained weights, does not outperform MiLoRA or CorDA in knowledge preservation }\cite{yang2024corda}.  This motivates us to turn our attention toward the approach of ensuring the space of adapters initialization is orthogonal to pre-trained knowledge.

From the perspective of making the space of LoRA initialization orthogonal to pre-trained knowledge, MiLoRA has done this by making the space of LoRA initialization orthogonal to $\mathbf{W}_0$. Given that the nonlinear neural network unit is $\sigma(\mathbf{W}\mathbf{x})$, we can also consider the null space of $\mathbf{X}_\text{pre}$. $\mathbf{X}_\text{pre}$ contains information from the parameters of all previous layers and the input data, whereas $\mathbf{W}_0$ only considers the current layer. This suggests that we should choose to analyze $\mathbf{X}_\text{pre}$. Meanwhile, we also show that the effective rank of $\mathbf{X}_{\text{pre}}$ is much smaller than that of $\mathbf{W}_0$. This means that a higher proportion of information in $\mathbf{X}_{\text{pre}}$ is concentrated in the subspace corresponding to the highest singular values, while the subspace associated with the smallest singular values contains a lower proportion of pre-trained knowledge. 

In this paper, we propose LoRA-Null, \textit{i.e.} Low-Rank Adaptation via Null Space of pre-trained knowledge activations, to address catastrophic forgetting. LoRA-Null randomly samples a small amount of data from datasets that are representative of the pre-training knowledge to get the activation $\mathbf{X}_\text{pre}$. Then, we extract the null space of $\mathbf{X}_\text{pre}$. Following \citeauthor{meng2024pissa,milora,yang2024corda}, we also let the component of $\mathbf{W}_0$ be the LoRA initialization. Thus, we make the LoRA initialization $\mathbf{B}\mathbf{A} = \mathbf{W}_0 \mathbf{U}_\text{null}\mathbf{U}_\text{null}^\top$. 

Our contributions can be concluded as follows:
\begin{itemize}
\item  We find that making the space of LoRA initialization orthogonal to pre-trained knowledge is more critical for preserving pre-trained knowledge than making the residual weights close to the pre-trained weights.

\item We find that the effective rank of the input activation is much smaller than that of the pre-trained weights. To push orthogonal LoRA initialization to the extreme for knowledge preservation, i.e., Low-Rank Adaptation via Null Space, which constructs adapters initialized from the null space of the pre-trained knowledge activation.

\item We conduct extensive experiments on three tasks, showing that our proposed LoRA-Null achieves good downstream performance while preserving pre-trained knowledge.

\end{itemize}

\section{Related Work}
\textbf{Initialization methods for LoRA}.
The vanilla LoRA uses a random Gaussian distribution to initialize the up-projection matrix $\mathbf{A}$ and uses a zero matrix to initialize the down-projection matrix $\mathbf{B}$. Recent studies have shown that the proper initialization can improve the performance of LoRA~\cite{meng2024pissa,yang2024corda,milora}. LoRA initialization methods can be broadly classified into two main categories: (1) those designed to accelerate downstream task performance, exemplified by PiSSA~\cite{meng2024pissa}, and the instruction-previewed mode of CorDA~\cite{yang2024corda}; and (2) those aimed at preserving pre-trained knowledge, such as MiLoRA~\cite{milora} and the knowledge-preserved mode of CorDA~\cite{yang2024corda}. PiSSA adopts principal singular values and singular vectors of the pre-trained weights as the initialization for LoRA adapters~\cite{meng2024pissa}. To extract input-relevant pre-trained weight components, CorDA performs SVD on the product of pre-trained weights and input activation covariance matrices. The top singular values/vectors are then multiplied by the inverse of the input activation's covariance matrix~\cite{yang2024corda}. CorDA operates in two distinct modes: an instruction-previewed mode, which utilizes input activations from the downstream task, and a knowledge-preserved mode, where input activations are sampled from the pre-training task. MiLoRA adopts minor singular values and singular vectors of the pre-trained weights as the initialization for LoRA adapters~\cite{milora}. In this paper, we focus on LoRA initialization methods designed for knowledge preservation and further elucidate the mechanisms of LoRA initialization in achieving knowledge preservation.\\
\textbf{Activation-aware knowledge preservation}. 
The activation is the output of a neuron after its weighted sum of inputs has been transformed by a non-linear activation function. Recent studies have shown that considering activation features can lead to better pre-trained knowledge preservation compared to only focusing on pre-trained weights~\cite{lin2024awq,wang2025svdllm,yang2024corda}. The activation-aware knowledge preservation is widely used in model compression and LoRA initialization. Activation-aware Weight Quantization (AWQ) is a model quantization method that identifies and scales a small percentage of crucial weights in LLMs based on activation distributions, significantly reducing quantization error and accelerating inference~\cite{lin2024awq}. SVD-LLM is an SVD-based post-training LLM compression method that uses truncation-aware data whitening derived from activations and sequential low-rank approximation to recover accuracy~\cite{wang2025svdllm}. CorDA extracts data-relevant pre-trained weight components by applying SVD to the product of pre-trained weights and input activation covariance matrices~\cite{yang2024corda}. The dataset sampled to construct the input activation is called the calibration set. To retain pre-training knowledge, these data are sampled from dataset that embodies the pre-training capabilities.

\section{Preliminaries}

\subsection{LoRA}
For pretrained weights $\mathbf{W}_0 \in \mathbb{R}^{d_\text{out} \times d_\text{in}}$, LoRA introduces a down-projection matrix $\mathbf{A} \in \mathbb{R}^{r \times d_\text{in}}$ and an up-projection matrix $\mathbf{B} \in \mathbb{R}^{d_\text{out} \times r}$, where $r \ll \min(d_\text{out}, d_\text{in})$, which is
\begin{equation}
\mathbf{y} = \mathbf{W}^{*} \mathbf{x} = \mathbf{W}_0 \mathbf{x} + \Delta \mathbf{W} \mathbf{x} = \mathbf{W}_0 \mathbf{x} + \mathbf{B} \mathbf{A} \mathbf{x},
\end{equation}
where $\mathbf{W}^{*}$ is the fine-tuned weights, $\Delta \mathbf{W}$ is the change in weights, $\mathbf{x} \in \mathbb{R}^{d_\text{in}}$ is the input, and $\mathbf{y} \in \mathbb{R}^{d_\text{out}}$ is the output.

\subsection{Input Activations of Pre-trained Knowledge: $\mathbf{X}_{\text{pre}}$}
Following~\citeauthor{yang2024corda}, we randomly collect some samples from the training data of some tasks that are representative of the pre-trained knowledge. Here, the collected samples are used to compute their corresponding input activations. Denote $\mathbf{X}_{\text{pre}} \in \mathcal{R}^{d_{\text{in}} \times BL}$ as the input activation matrix of a linear layer derived from these sampled data, where $d_{\text{in}}$ is the input dimension, $B$ is the number of collected samples, and $L$ represents the maximum sequence length. \textbf{Note that $\mathbf{X}_{\text{pre}}$ denotes the input activations of pre-trained knowledge, while $\mathbf{x}$ denotes any input}.

\subsection{MiLoRA and CorDA}
\label{sec:milora_and_corda}
MiLoRA \cite{milora} only updates the minor singular components of the weight matrix while freezing the principal singular components. Let $ \verb|SVD|(\mathbf{W}_0) = \mathbf{\hat{U}} \mathbf{\hat{\Sigma}} \mathbf{\hat{V}}^\top $ and $R = \text{rank}(\mathbf{W}_0)$, MiLoRA uses $ \mathbf{\hat{U}}_{[:,:R-r]} \mathbf{\hat{\Sigma}}_{[:R-r]} \mathbf{\hat{V}}_{[:R-r,:]}^\top $ as the residual weight and use $\mathbf{\hat{U}}_{[,-r:]} \mathbf{\hat{\Sigma}}_{[-r:]} \mathbf{\hat{V}}^\top_{[-r:,:]} $ as the LoRA initialization. CorDA \cite{yang2024corda} is a LoRA initialization method that uses data from downstream tasks or pre-trained knowledge. CorDA uses the covariance matrix $\mathbf{C} = \mathbf{X}_{\text{pre} }\mathbf{X}_{\text{pre}}^\top$ to extract the components of the pre-trained weight matrices that are the most related to the provided data. CorDA first uses $\verb|SVD|(\mathbf{W}_0\mathbf{C}) = \mathbf{\hat{U}} \mathbf{\hat{\Sigma}} \mathbf{\hat{V}}^\top$. For the knowledge-preserved adaptation of CorDA, CorDA uses $\mathbf{\mathbf{\hat{U}}}_{[:,:R-r]} \mathbf{\mathbf{\hat{\Sigma}}}_{[:R-r]} \mathbf{\mathbf{\hat{V}}}^\top_{[:R-r,:]} \mathbf{C}^{-1}$ as the residual weights and uses $\mathbf{\mathbf{\hat{U}}}_{[:,-r:]} \mathbf{\mathbf{\hat{\Sigma}}}_{[-r:]} \mathbf{\mathbf{\hat{V}}}^{\top}_{[-r:,:]} \mathbf{C}^{-1}$.

\subsection{Effective Rank}
Given a non-zero matrix $\mathbf{K} \in \mathbb{R}^{M \times N}$ with singular values $\sigma_1, \sigma_2, \dots, \sigma_Q$ where $Q = \min\{M,N\}$, we define the normalized singular value distribution:

\begin{equation}
p_i =  \frac{\sigma_i}{\sum_{j=1}^Q \sigma_j}, \quad i = 1,\ldots,Q,
\end{equation}

\noindent where $p_i$ represents the relative weight of the $i$-th singular value in the spectrum. The \textbf{effective rank} is defined as

\begin{equation}
\mathrm{eRank}(\mathbf{K}) = \exp\left( -\sum_{i=1}^{Q} p_i \log p_i \right),
\label{erank}
\end{equation}

\noindent where it quantifies the spectral entropy of singular values, giving an information-theoretic characterization of a matrix's effective rank \cite{erank}. A high effective rank indicates that a matrix’s singular values are relatively evenly distributed, while a low effective rank indicates that most information is concentrated in the top few singular values.

\section{Motivation}
\subsection{The analysis of CorDA and MiLoRA from the perspective of $\mathbf{W}'_0$}
In this section, we first analyze the CorDA and MiLoRA from the perspective of $\mathbf{W}'_0$. We present two theorems as follows, where the proofs are given in extended version.
\begin{theorem}
The initialization of MiLoRA is the solution of 
\begin{align}
\begin{aligned}
    \min_{\mathbf{W}_0'}  \|\mathbf{W}_0' - \mathbf{W}_0\|_{\mathrm{F}}, \text{s.t.} \;  \text{rank} \left( \mathbf{W}_0' \right) = R - r .
    \label{eq:formulation1}
\end{aligned}
\end{align}
\label{theorem1}
\end{theorem}

\begin{theorem}
The initialization of CorDA is the solution of 
\begin{align}
\begin{aligned}
 \min_{\mathbf{W}_0'}  \|\mathbf{W}_0'\mathbf{X}_{\text{pre}} - \mathbf{W}_0\mathbf{X}_{\text{pre}}\|_{\mathrm{F}}, \text{s.t.} \; \text{rank} \left( \mathbf{W}_0' \right) = R - r.
    \label{eq:formulation2}
\end{aligned}
\end{align}
\label{theorem2}
\end{theorem} 

We can observe that both MiLoRA and CorDA aim to make $\mathbf{W}'_0 $  close to $\mathbf{W}_0$. However, in Figure \ref{fig:llama3.2_relative}, we have shown that the contribution of $\mathbf{B}\mathbf{A}$ after fine-tuning to the pre-trained knowledge is non-negligible. Meanwhile, we note that although LoRA freezes the original weights $\mathbf{W}_0$, it does not preserve the pre-trained knowledge well. Therefore, it is not so important to make $\mathbf{W}'_0$ close to $\mathbf{W}_0$.

\begin{figure}[t!]
    \centering
    \begin{subfigure}{0.23\textwidth}
        \centering
        \includegraphics[width=\linewidth]{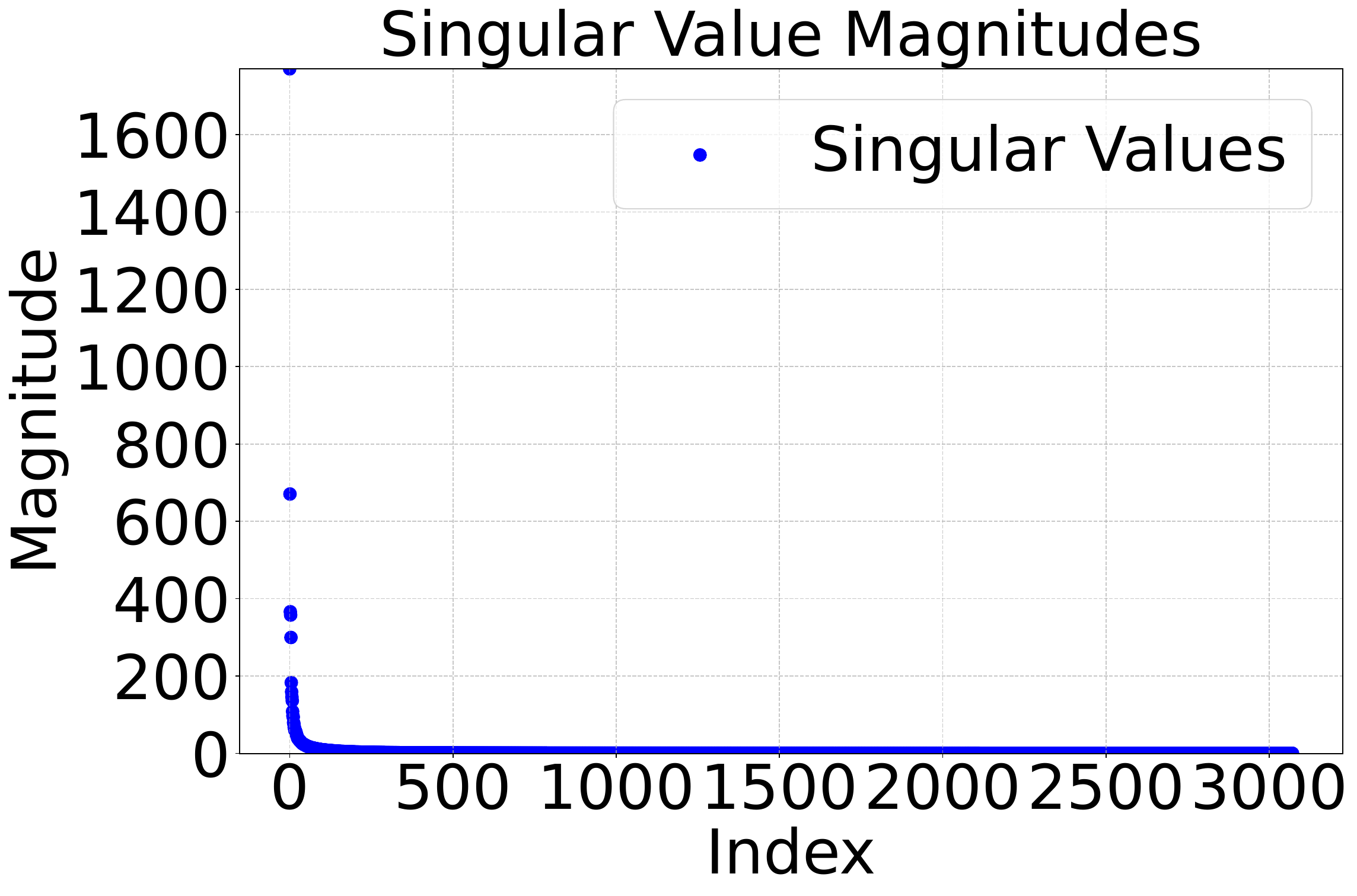}
        \caption{$\mathbf{X}_\text{pre}$}
        \label{fig:llama3.2_0_k_erank_Xpre}
    \end{subfigure}%
    %\hfill
    \begin{subfigure}{0.23\textwidth}
        \centering
        \includegraphics[width=\linewidth]{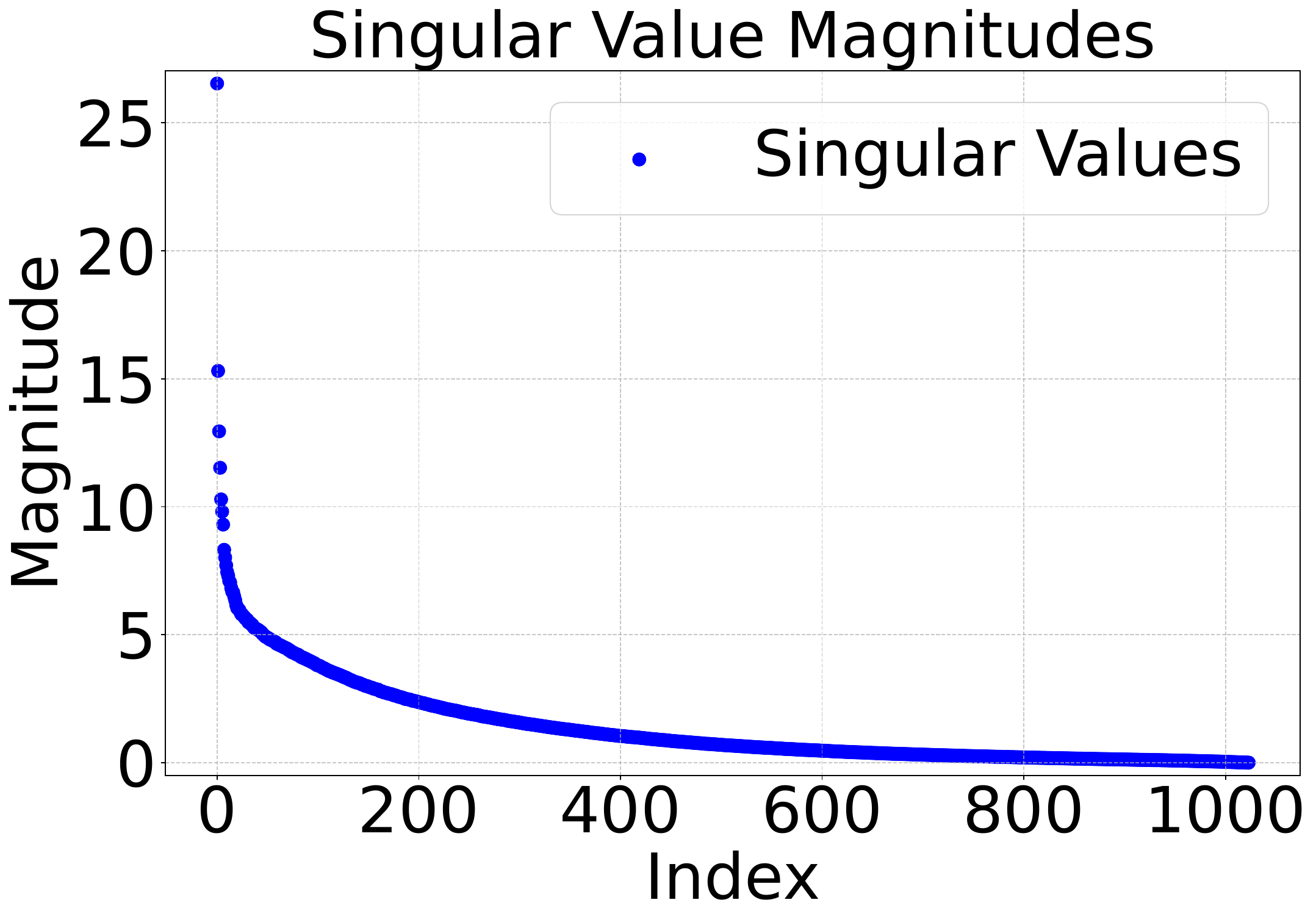}
        \caption{$\mathbf{W}_0$}
        \label{fig:llama3.2_0_k_erank_W0}
    \end{subfigure}%

    \caption{The singular value magnitudes of $\mathbf{X}_\text{pre}$ and $\mathbf{W}_0$ on the key matrix of layer 0 on LLaMA-3.2-3B}
    \label{fig:llama3.2_singular_values}
\end{figure}

\begin{table}[t!]
\centering
\small % ✅ 确保 10pt Roman 字体

\setlength{\tabcolsep}{1mm} % ✅ 关键：减小列间距
\begin{tabular}{l|c|c|c|c|c}
\toprule
 & qproj & kproj & vproj  & upproj & downproj \\
\midrule
$\mathbf{W}_0$ (0) & 1214.11 & 548.30 & 892.36   & 2871.72 & 2883.30 \\
\midrule
$\mathbf{X}_{\text{pre}}$ (0) & 101.28 & 101.28 & 101.28   & 553.59 & 758.06 \\
\midrule
$\mathbf{W}_0$ (1) & 1788.42 & 714.95 & 941.63 & 2858.23 & 2882.98 \\
\midrule
$\mathbf{X}_{\text{pre}}$ (1) & 85.63 & 85.63 & 85.63  & 657.79 & 1.27 \\
\midrule
$\mathbf{W}_0$ (27) & 2089.48 & 851.92 & 943.69  & 2757.13 & 2838.64 \\
\midrule
$\mathbf{X}_{\text{pre}}$ (27) & 194.22 & 194.22 & 194.22  & 302.78 & 79.97 \\
\bottomrule
\end{tabular}

\caption{The effective rank of $\mathbf{X}_{\text{pre}}$ and $\mathbf{W}_0$ on LLaMA-3.2-3B. The number in ``()'' indicates the layer index.}
\label{tab:effecitve_ranks}
\end{table}

\subsection{The analysis of CorDA and MiLoRA from the perspective of the LoRA initialization space}
From the perspective of the LoRA initialization space, we observe that the initialization space of MiLoRA is the null space of $\mathbf{W}_0$, whereas the initialization space of CorDA is neither the null space of $\mathbf{W}_0$ nor that of $\mathbf{X}_{\text{pre}}$.

When the LoRA initialization space contains the principal components of pre-trained knowledge, since the LoRA matrices are trainable, LoRA may drastically amplify the components associated with pre-trained knowledge; moreover, these principal components, when perturbed, can more easily lead to the corruption of the pre-trained knowledge. \textbf{Therefore, it is the key for knowledge preserving to make $\mathbf{B}\mathbf{A}$ \text{orthogonal to pre-trained knowledge}.} 

\subsection{Comparing the null space of $ \mathbf{X}_{\text{pre}} $ and $\mathbf{W}_0 $}

The output of a neural network is determined by a series of nonlinear units \(\sigma(\mathbf{W} \mathbf{x})\). Thus, for the null space for LoRA initialization, we can take into account either $\mathbf{W}_0$ or $ \mathbf{X}_{\text{pre}}$. Recent studies have shown that $\mathbf{X}_{\text{pre}}$ can reflect the characteristics of pre-trained data \cite{lin2024awq,wang2025svdllm,yang2024corda}. Next, we will explain why the null space of $\mathbf{X}_{\text{pre}}$ is better than that of $ \mathbf{W}_0$, from two perspectives.

First, $\mathbf{W}_0$ has weaker importance of information contained in pre-training tasks than $\mathbf{X}_\text{pre}$. $\mathbf{X}_{\text{pre}}$ takes into account the parameters from all previous layers as well as the input data, whereas $\mathbf{W}_0$ only reflects its own parameters.

Second, we analyze $\mathbf{X}_{\text{pre}}$ and $\mathbf{W}_0$ from the perspective of effective rank. The effective rank is calculated by Eq.\ref{erank}. In Figure \ref{fig:llama3.2_singular_values}, we can observe that the singular value magnitudes of $\mathbf{X}_{\text{pre}}$ are steeper than those of $\mathbf{W}_0$. This means that the information contained in $\mathbf{X}_{\text{pre}}$ is more concentrated in the principal subspace compared to $\mathbf{W}_0$. From the perspective of effective rank, this implies that the effective rank of $\mathbf{X}_{\text{pre}}$ is much smaller than that of $\mathbf{W}_0$, as is demonstrated in Table~\ref{tab:effecitve_ranks}. Actually, $\mathbf{X}_{\text{pre}}$ contains information about all the parameters and input data from previous layers, which is the reason for its smaller effective rank. This is because the more precise the information in $\mathbf{X}_{\text{pre}}$ is, the lower the uncertainty of the information it contains, and thus the lower the effective rank will be. From the perspective of effective rank alone, the null space of $\mathbf{X}_{\text{pre}}$ contains less pre-trained knowledge compared to that of $\mathbf{W}_0$. Therefore, initializing LoRA in the null space of $\mathbf{X}_{\text{pre}}$ will achieve better knowledge preservation than initializing it in the null space of $\mathbf{W}_0$.

\begin{figure}[t!]
	\centering
    \includegraphics[width=\linewidth]{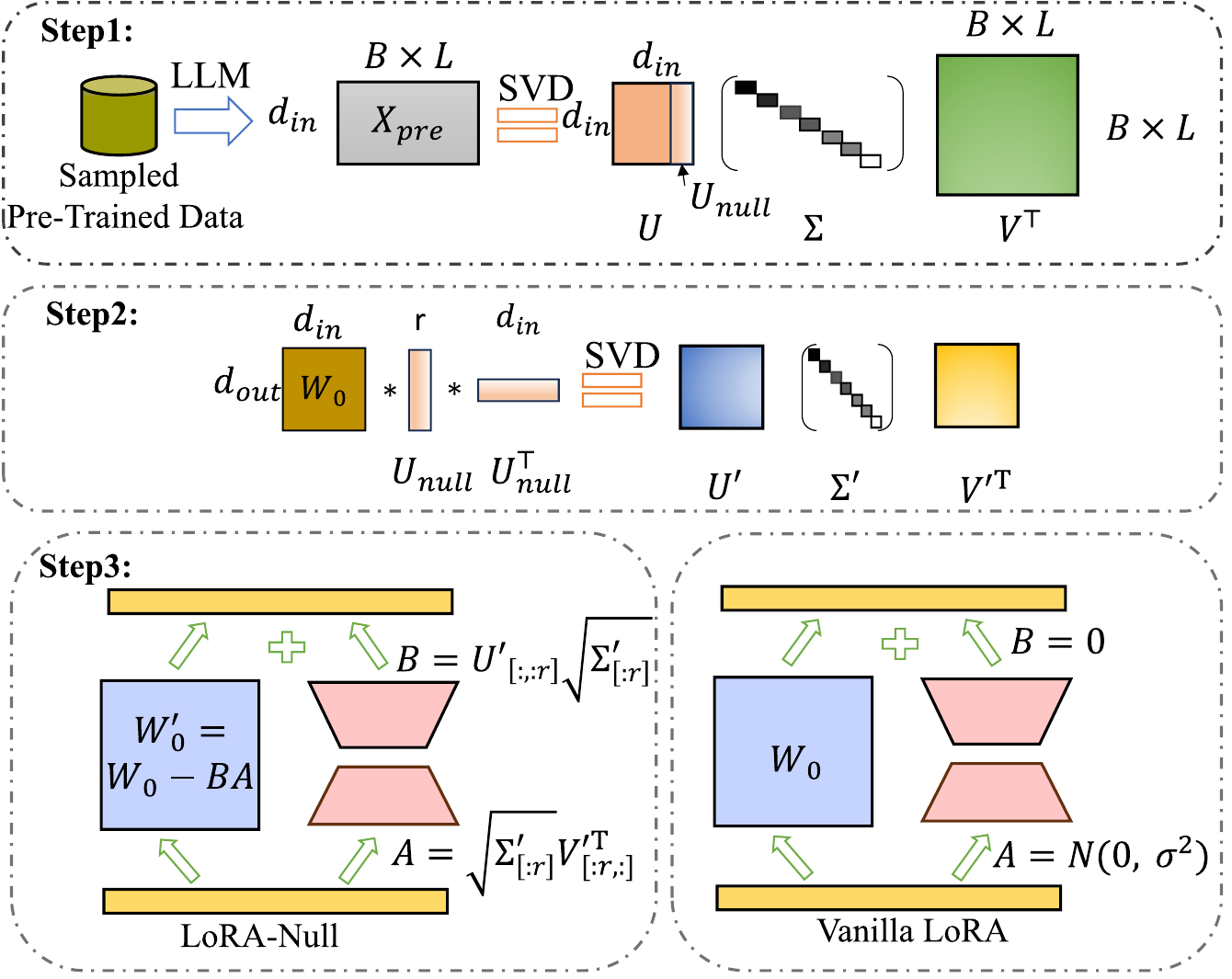}
	\caption{ An illustration of LoRA-Null. We first use SVD on the pre-trained knowledge activations to obtain the $\mathbf{U}_{\text{null}}$. Then, we use $\mathbf{W}_0 (\mathbf{U}_{\text{null}} \mathbf{U}_{\text{null}}^\top)$ to extract the projection of $\mathbf{W}_0$ onto the null space of $\mathbf{X}_{\text{pre}}$ . Finally, we conduct SVD on $\mathbf{W}_0(\mathbf{U}_{\text{null}} \mathbf{U}_{\text{null}}^\top)$ to initialize $\mathbf{A}$ and $\mathbf{B}$ and replace $\mathbf{W}_0$ with $\mathbf{W}_0 - \mathbf{B} \mathbf{A}$. \textbf{LoRA-Null} only allows updating $\mathbf{A}$ and $\mathbf{B}$.}
	\label{fig:lora_null}
\end{figure}

\begin{table*}[t!]
\centering
\small

\begin{subtable}{\linewidth}
\centering
\subcaption{Math on LLaMA-2-7B}
\label{llama2_math}
\setlength{\tabcolsep}{1mm}
\begin{tabular}{l|c|ccccc|ccc|c}
    \toprule[1.5pt]
    Method & \textbf{\#Para} & Trivia QA & NQ open & WebQS & Avg1 & Avg1(Per) & GSM8k & Math & Avg2 & GM \\
    \midrule
    LLaMA-2-7B & - & 52.51 & 18.83 & 5.91 & 25.75 & 100.00 & - & - & - & - \\
    \midrule
    LoRA & 320M & 45.95 & 1.16 & \textbf{6.64} & 17.91 & 64.56 & 42.99 & 6.26 & 24.63 & 21.00 \\
    PiSSA & 320M & 43.70 & 2.30 & 6.50 & 17.50 & 65.15 & \textbf{51.70} & 7.64 & \textbf{29.67} & \underline{22.79} \\
    MiLoRA & 320M & 47.02 & 3.66 & 6.10 & 18.93 & 69.66 & 41.47 & 6.20 & 23.84 & 21.24 \\
    CorDA & 320M & \underline{48.99} & \underline{7.15} & 5.76 & \underline{20.63} & \underline{76.24} & 41.47 & \underline{8.22} & 24.85 & 22.64 \\
    LoRA-Null & 320M & \textbf{50.02} & \textbf{7.98} & \underline{6.55} & \textbf{21.52} & \textbf{79.21} & \underline{44.43} & \textbf{8.80} & \underline{26.62} & \textbf{23.93} \\
    \bottomrule[1.5pt]
\end{tabular}
\end{subtable}

\begin{subtable}{\linewidth}
\centering
\subcaption{Code on LLaMA-2-7B}
\label{llama2_code}
\setlength{\tabcolsep}{1mm}
\begin{tabular}{l|c|ccccc|ccc|c}
    \toprule[1.5pt]
    Method & \textbf{\#Para} & Trivia QA & NQ open & WebQS & Avg1 & Avg1(per) & HumanEval & MBPP & Avg2 & GM \\
    \midrule
    LLaMA-2-7B & - & 52.51 & 18.83 & 5.91 & 25.75 & 100.00 & - & - & - & - \\
    \midrule
    LoRA & 320M & \textbf{51.30} & 12.24 & \underline{8.66} & \underline{24.07} & 87.57 & 17.19 & 21.29 & 19.24 & 21.52 \\
    PiSSA & 320M & 46.08 & 14.40 & \textbf{9.25} & 23.24 & 84.25 & \textbf{20.61} & \textbf{23.84} & \textbf{22.23} & \textbf{22.73} \\
    MiLoRA & 320M & 49.33 & 13.49 & 7.19 & 23.34 & 88.53 & \underline{18.36} & 20.97 & 19.67 & 21.43 \\
    CorDA & 320M & 49.01 & \underline{15.24} & 6.59 & 23.61 & \underline{91.42} & \underline{18.36} & \underline{21.65} & \underline{20.01} & 21.74 \\
    LoRA-Null & 320M & \underline{51.29} & \textbf{16.51} & 7.87 & \textbf{25.22} & \textbf{95.12} & \underline{18.36} & \underline{21.65} & \underline{20.01} & \underline{22.46} \\
    \bottomrule[1.5pt]
\end{tabular}
\end{subtable}

\begin{subtable}{\linewidth}
\centering
\subcaption{Instruction Following on LLaMA-2-7B}
\label{llama2_MT}
\setlength{\tabcolsep}{1mm}
\begin{tabular}{l|c|ccccc|c|c}
    \toprule[1.5pt]
    Method & \textbf{\#Param} & Trivia QA & NQ open & WebQS & Avg1 & Avg1 (Per) & MTBench & GM \\
    \midrule
    LLaMA-2-7B & - & 52.51 & 18.83 & 5.91 & 25.75 & 100.00 & - & - \\
    \midrule
    LoRA & 320M & 46.06 & 9.72 & \textbf{7.28} & 21.02 & 79.78 & 3.81 & 8.95 \\
    PiSSA & 320M & 35.76 & 10.17 & 5.61 & 17.18 & 72.34 & \textbf{4.21} & 8.50 \\
    MiLoRA & 320M & 45.98 & 11.94 & \underline{6.74} & 21.55 & 83.66 & 3.79 & 9.04 \\
    CorDA & 320M & \underline{47.10} & \underline{13.63} & \underline{6.74} & \underline{22.49} & \underline{87.36} & 3.89 & \underline{9.35} \\
    LoRA-Null & 320M & \textbf{48.09} & \textbf{14.71} & 6.69 & \textbf{23.16} & \textbf{89.90} & \underline{4.02} & \textbf{9.65} \\
    \bottomrule[1.5pt]
\end{tabular}
\end{subtable}

\begin{subtable}{\linewidth}
\centering
\subcaption{Math on LLaMA3.2-3B}
\label{llama3.2_math}
\setlength{\tabcolsep}{1mm}
\begin{tabular}{l|c|ccccc|ccc|c}
    \toprule[1.5pt]
    Method & \textbf{\#Para} & Trivia QA & NQ open & WebQS & Avg1 & Avg1(Per) & GSM8k & Math & Avg2 & GM \\
    \midrule
    LLaMA-3.2-3B & - & 50.77 & 13.55 & 9.25 & 24.52 & 100.00 & - & - & - & - \\
    \midrule
    LoRA & 195M & 43.62 & 8.28 & 8.02 & 19.97 & 77.91 & 55.65 & 12.72 & 34.19 & 26.13 \\
    PiSSA & 195M & 42.88 & 7.26 & 8.91 & 19.68 & 78.12 & \textbf{63.84} & \textbf{15.7} & \textbf{39.77} & 27.98 \\
    MiLoRA & 195M & 46.26 & 10.97 & 8.17 & 21.80 & 86.80 & 56.56 & 13.58 & 35.07 & 27.65 \\
    CorDA & 195M & 46.77 & 10.22 & \textbf{9.20} & \underline{22.06} & \underline{89.00} & \underline{58.91} & \underline{14.70} & \underline{36.81} & \underline{28.50} \\
    LoRA-Null & 195M & \textbf{49.03} & \textbf{11.52} & \underline{9.01} & \textbf{23.19} & \textbf{93.18} & 58.76 & 14.06 & 36.41 & \textbf{29.06} \\
    \bottomrule[1.5pt]
\end{tabular}
\end{subtable}

\caption{Results of LoRA-Null vs. baselines on LLaMA-2-7B (LLaMA-3.2-3B). The first row shows the pre-trained performance. \textbf{Bold} indicates the best result; \underline{underlined} indicates the runner-up. ``Avg1'' is the average preservation score, ``Avg1(Per)'' is the percentage of preserved performance relative to pre-trained, ``Avg2'' is the average of downstream task scores, and ``GM'' is the geometric mean of Avg1 and Avg2.}
\label{llama2_result}
\end{table*}

\section{Our Method}

In this section, we introduce our proposed LoRA-Null. We let $\mathbf{B}\mathbf{A}$ in the null space of $\mathbf{X}_{\text{pre}}$ and do not consider the residual weight $\mathbf{W}'_0$. 
Following MiLoRA and CorDA \cite{yang2024corda,milora}, we let $\mathbf{B}\mathbf{A}$ is the component of $\mathbf{W}_0$. Thus, we let $\mathbf{B}\mathbf{A} = \mathbf{W}_0 \mathbf{U}_\text{null}\mathbf{U}_\text{null}^\top$, where $\mathbf{U}_\text{null}$ is the left null space of $\mathbf{X}_{\text{pre}}$. $\mathbf{B}\mathbf{A} = \mathbf{W}_0 \mathbf{U}_\text{null}\mathbf{U}_\text{null}^\top$ means $\mathbf{B}\mathbf{A}$ is the projection of $\mathbf{W}_0$ onto the null space of $\mathbf{X}_{\text{pre}}$. In contrast, the LoRA initialization of MiLoRA is the projection of $\mathbf{W}_0$ onto the null space of $\mathbf{W}_0$.

To get the left null space of $\mathbf{X}_{\text{pre}}$, we perform  SVD of $\mathbf{X}_{\text{pre}}$:
\begin{equation}
    \mathbf{X}_{\text{pre}} = \mathbf{U} \mathbf{\Sigma} \mathbf{V}^\top = \sum_{i=1}^{R} \sigma_i \mathbf{u}_i \mathbf{v}_i^\top,
    \label{SVD_OF_Xpre}
\end{equation}
where $\mathbf{U} \in \mathbb{R}^{d_{\text{in}} \times d_{\text{in}}}$ and $\mathbf{V} \in \mathbb{R}^{d_{B \times L} \times d_{B \times L}}$ are orthogonal matrices, and $\mathbf{\Sigma} \in \mathbb{R}^{d_{\text{in}} \times d_{B \times L}}$ is a diagonal matrix with singular values $\sigma_1 \geq \sigma_2 \geq \dots \geq \sigma_R \geq 0$ (with $R = \min (d_{\text{in}}, B\times L)=d_{\text{in}}$, $B \times L \gg d_{\text{in}}$ ). Given a predefined LoRA rank $r$ (typically $r \ll d_{\text{in}}$), we approximate the left null space of $\mathbf{X}_{\text{pre}}$ by selecting the $r$ left singular vectors associated with the \textit{smallest} singular values. Formally, if $\mathbf{U} = [\mathbf{U}_1 | \mathbf{U}_2]$ with $\mathbf{U}_2 \in \mathbb{R}^{d_{\text{in}} \times r}$, we treat $\mathbf{U}_2$ as the approximate null space under the assumption that the trailing singular values $\{\sigma_{d_{\text{in}}-r+1}, \dots, \sigma_{d_{\text{in}}}\}$ are negligibly small, as evidenced in Figure \ref{fig:llama3.2_singular_values}. This leads to the approximation:
\begin{equation}
    \mathbf{U}_2^\top \mathbf{X}_{\text{pre}} \approx \mathbf{0}.
\end{equation}
Hereafter, we use $\mathbf{U}_{\text{null}}$ to represent $\mathbf{U}_2$.

As shown in Figure \ref{fig:lora_null}, we perform SVD on the pre-trained weights projected onto the null space, which is:
\begin{equation}
    \mathbf{W}_0 \mathbf{U}_{\text{null}} \mathbf{U}_{\text{null}}^\top = \mathbf{U}' \mathbf{\Sigma}' \left(\mathbf{V}'\right)^\top,
\end{equation}
where $\mathbf{U}'$ and $\mathbf{V}'$ are orthogonal matrices of left and right singular vectors, respectively, and $\mathbf{\Sigma}'$ is a diagonal matrix of singular values. And we can observe that 
\begin{equation}
    \text{rank}\left(\mathbf{W}_0 \mathbf{U}_{\text{null}} \mathbf{U}_{\text{null}}^\top\right) 
    \leq \text{rank}\left(\mathbf{U}_{\text{null}} \right) = r.
\end{equation}
We then initialize the adapter matrices $\mathbf{B}$ and $\mathbf{A}$ as:

\begin{align}
    \mathbf{B} = \mathbf{U}_{[:,:r]}' \sqrt{\mathbf{\Sigma}'_{[:r]}},  \mathbf{A} = \sqrt{\mathbf{\Sigma}'_{[:r]}} \mathbf{V}_{[:r,:]}'^\top.
\end{align}

We prove that the columns of $\mathbf{A}^\top$ lie in the column space of $\mathbf{U}_{\text{null}}$ in extended version. The residual weight matrix is
\begin{equation}
    \mathbf{W}'_0 = \mathbf{W}_0 - \mathbf{B}\mathbf{A}.
\end{equation}

During fine-tuning, only $\mathbf{A}$ and $\mathbf{B}$ are updated while $\mathbf{W'}_0$ are frozen. This is our proposed LoRA-Null.

\begin{theorem}%[3.1]
The initialization of LoRA-Null is not the solution of 
Eq. \ref{eq:formulation1} and Eq. \ref{eq:formulation2}.
\label{theorem:3}
\end{theorem}

The proof of Theorem \ref{theorem:3} and the corresponding experiments are provided in extended version. 
Theorem \ref{theorem:3} shows that LoRA-Null only considers the space of LoRA initialization and does not consider making $\mathbf{W}'_0 $  close to $\mathbf{W}_0$. This is the origin of our title: we push the idea of aligning the space of LoRA initialization to be orthogonal to pre-trained knowledge to the extreme in order to preserve pre-trained knowledge without considering making $\mathbf{W}'_0$ close to $\mathbf{W}_0$.

\section{Experiments}
\subsection{Experimental Setup}
\textbf{Models and Datasets}. We fine-tune the pre-trained LLMs---\textbf{LLaMA-2-7B} and \textbf{LLaMA-3.2-3B}---on \textbf{Math}, \textbf{Code}, and \textbf{Instruction Following} tasks. Following \citeauthor{yang2024corda}, the pre-trained knowledge is assessed using exact match scores (\%) on the \textbf{TriviaQA} \cite{triviaqa}, \textbf{NQ Open} \cite{nq_open}, and \textbf{WebQS} \cite{webqs} datasets. For Math, LLMs are trained on MetaMathQA \cite{yu2024metamath} and tested on the \textbf{GSM8k} \cite{cobbe2021training} and \textbf{Math} \cite{yu2024metamath} validation sets. For Code, LLMs are trained on CodeFeedback \cite{pmlr-v235-wei24h} and tested on the \textbf{Humaneval} \cite{humaneval} and \textbf{MBPP} \cite{mbpp}. For Instruction Following, LLMs are trained on WizardLM-Evol-Instruct \cite{xu2024wizardlm} and tested on the \textbf{MTBench} \cite{mtbench}. \\
\textbf{Implementation Details}.
 We follow the same training configuration as \cite{meng2024pissa,yang2024corda}. Specifically, the optimization process utilizes the AdamW \cite{loshchilov2018decoupled} optimizer with a batch size of 128 and a learning rate of $2 \times 10^{-5}$. We use cosine annealing schedules alongside a warmup ratio of 0.03. The training is conducted exclusively on the initial 100,000 conversations from the dataset for one epoch, where the loss calculation is based solely on the response. The rank of LoRA is set to 128. Our experiments are carried out on a single NVIDIA H800 80GB GPU. Following \cite{yang2024corda}, we randomly sample 256 data points from NQ Open with a maximum length of 1024 as the calibration set, which is used to construct the input activations representing the pre-trained knowledge. \\ %We use the same sample random seed as \cite{yang2024corda}.\\
\textbf{Baselines}. The baselines are: (1) LoRA~\cite{hu2022lora}; (2) PiSSA~\cite{meng2024pissa}; (3) MiLoRA~\cite{milora}; and (4) CorDA~\cite{yang2024corda}.

\begin{figure*}[t]
    \centering

    \begin{subfigure}{0.31\textwidth}
        \centering
        \includegraphics[width=\linewidth]{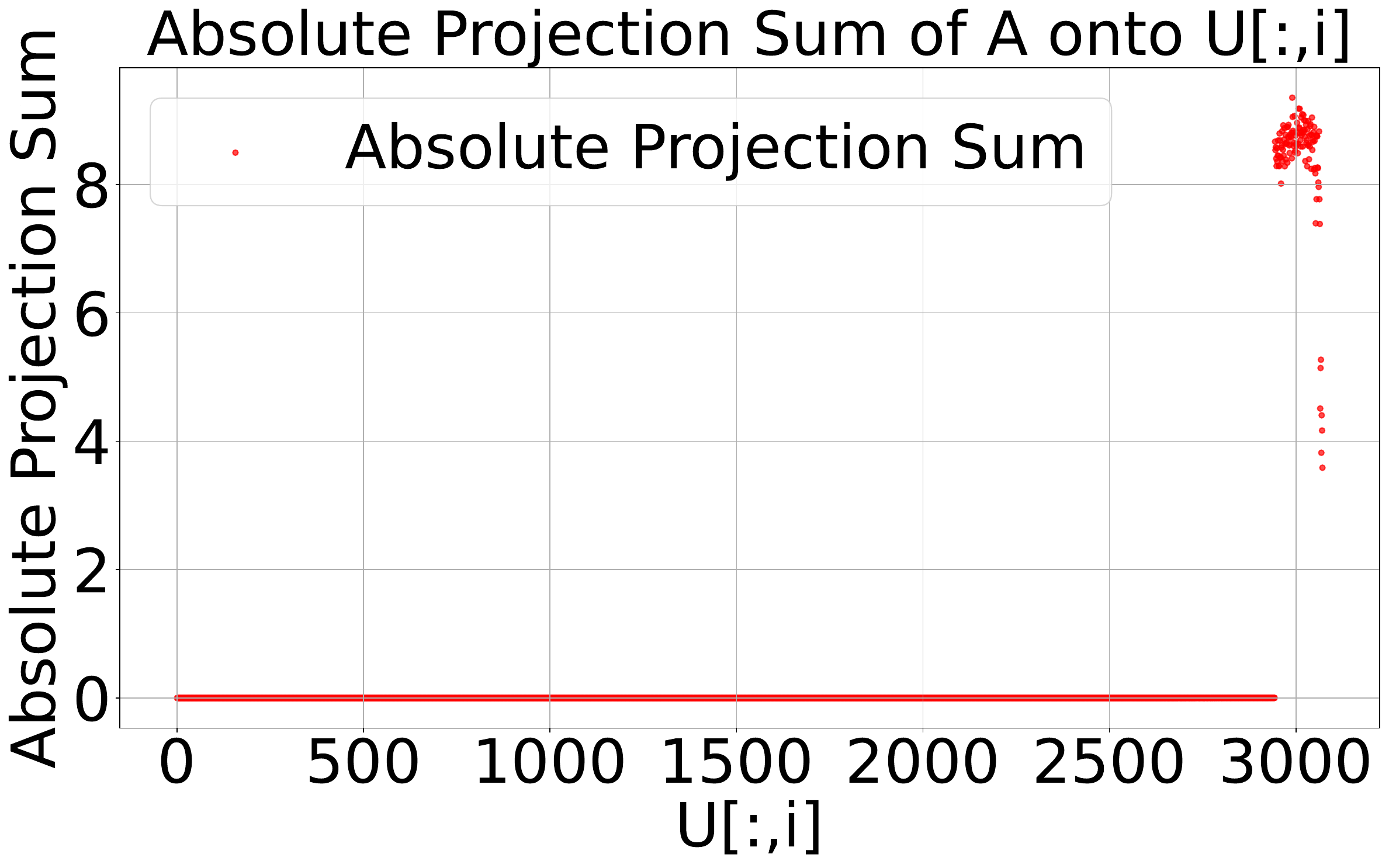}
        \caption{LoRA-Null}
        \label{fig:llama3.2_k_lora_null}
    \end{subfigure}%
    \hfill
    \begin{subfigure}{0.31\textwidth}
        \centering
        \includegraphics[width=\linewidth]{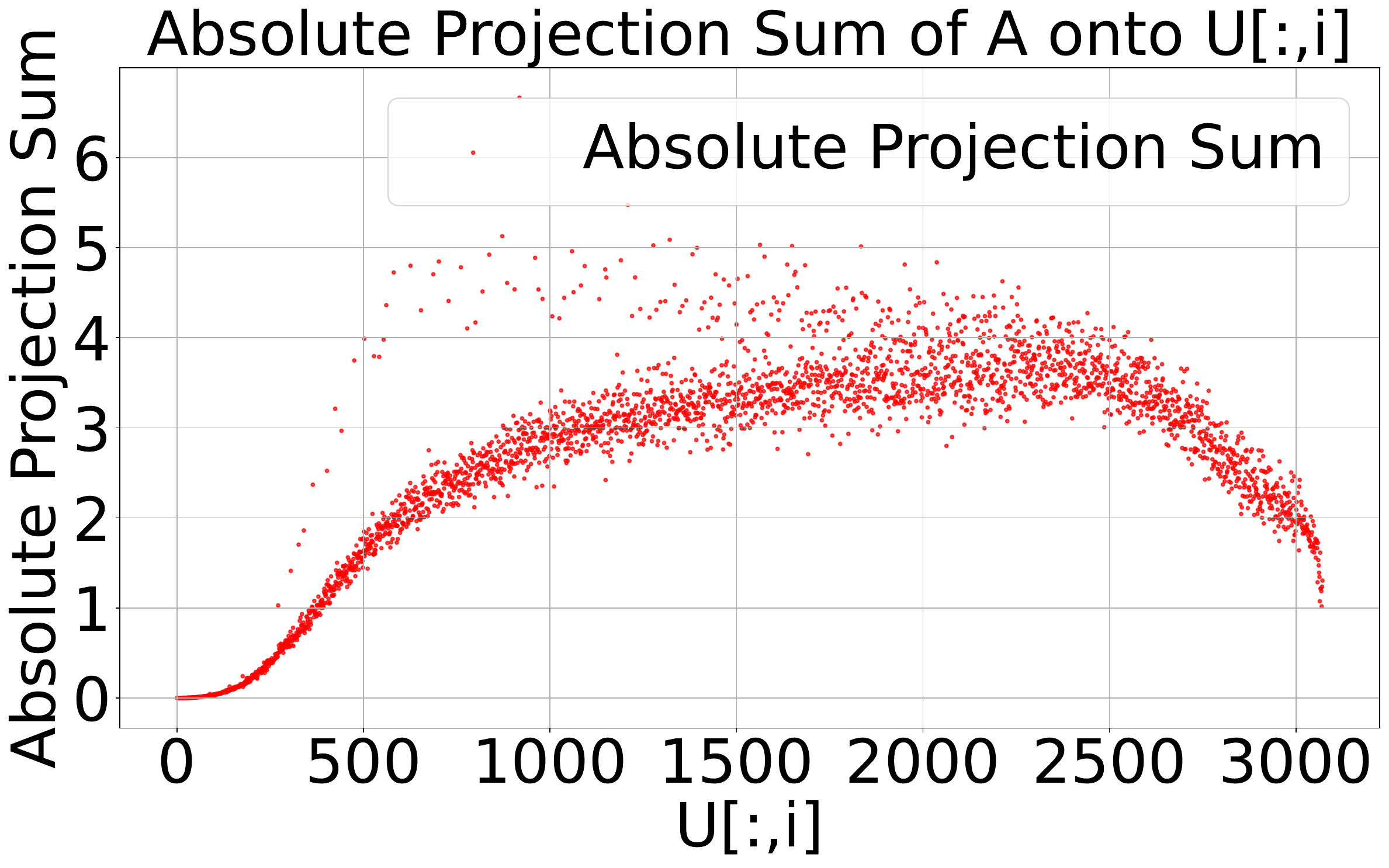}
        \caption{CorDA}
        \label{fig:llama3.2_k_corda}
    \end{subfigure}%
    \hfill
    \begin{subfigure}{0.31\textwidth}
        \centering
        \includegraphics[width=\linewidth]{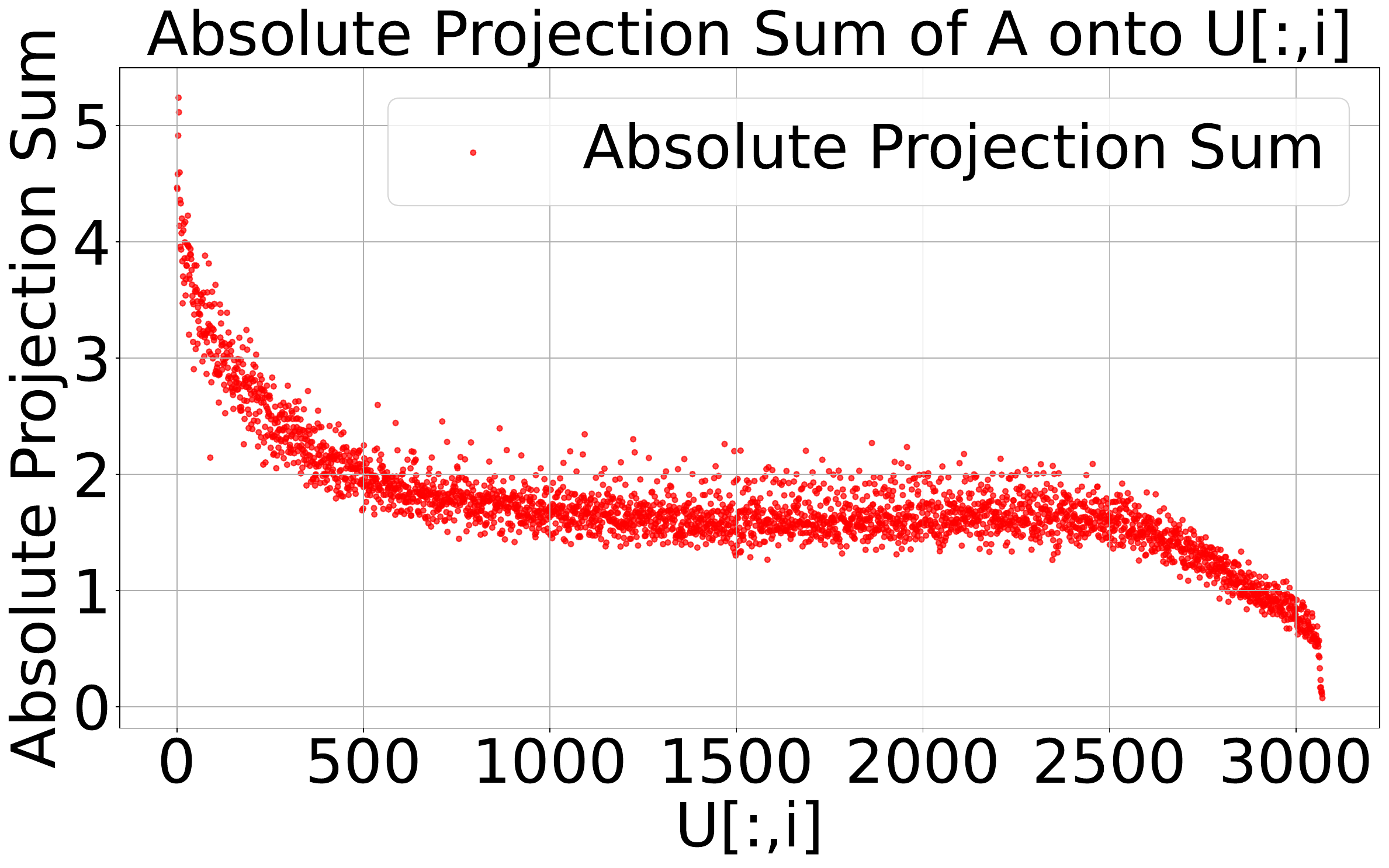}
        \caption{MiLoRA}
        \label{fig:llama3.2_k_milora}
    \end{subfigure}

    \caption{Component of the down-projection matrix $ \mathbf{A} $ onto the subspace of $ \mathbf{U}_{[:,i]} $ for the key matrices in LoRA-Null, CorDA and MiLoRA, applied to layer 0 of LLaMA-3.2-3B.}
    \label{fig:llama3.2_intro2}
\end{figure*}

\begin{table*}[t!]
\centering
\small

\begin{subtable}{\linewidth}
\centering
\subcaption{Different number of calibration sets}
\label{llama3.2_ab_sample}
\setlength{\tabcolsep}{1mm} 
\begin{tabular}{l|c|ccccc|ccc|c}
    \toprule[1.5pt]
    Method & \textbf{\#Rank Size, \#Samples Size} & Trivia QA & NQ open & WebQS & Avg1 & Avg1(Per) & GSM8k & Math & Avg2 & GM \\
    \midrule
    LLaMA-3.2-3B & - & 50.77 & 13.55 & 9.25 & 24.52 & 100.00 & - & - & - & - \\
    \midrule
    CorDA  & 128, 64 & 42.72 & 10.00 & 4.53 & 19.08 & 68.97 & 62.32 & 15.40 & 38.86 & 27.23 \\
    LoRA-Null  & 128, 64 & 48.69 & 12.33 & 9.06 & 23.36 & 94.95 & 59.59 & 13.94 & 36.77 & 29.31 \\
    \midrule
    CorDA  & 128, 256 & 46.77 & 10.22 & 9.20 & 22.06 & 89.00 & 58.91 & 14.70 & 36.81 & 28.50 \\
    LoRA-Null  & 128, 256 & 49.03 & 11.52 & 9.01 & 23.19 & 93.00 & 58.76 & 14.06 & 36.41 & 29.06 \\
    \midrule
    CorDA  & 128, 1024 & 48.46 & 9.89 & 9.30 & 22.55 & 89.48 & 58.15 & 14.50 & 36.33 & 28.62 \\
    LoRA-Null  & 128, 1024 & 48.98 & 11.14 & 10.78 & 23.63 & 92.90 & 59.59 & 14.54 & 37.07 & 29.60 \\
    \bottomrule[1.5pt]
\end{tabular}
\end{subtable}

\begin{subtable}{\linewidth}
\centering
\subcaption{Different ranks of LoRA adapters}
\label{llama3.2_ab_rank}
\setlength{\tabcolsep}{1mm} % ✅ 适当减小列间距
\begin{tabular}{l|c|ccccc|ccc|c}
    \toprule[1.5pt]
    Method & \textbf{\#Rank Size, \#Samples Size} & Trivia QA & NQ open & WebQS & Avg1 & Avg1(Per) & GSM8k & Math & Avg2 & GM \\
    \midrule
    LLaMA-3.2-3B & - & 50.77 & 13.55 & 9.25 & 24.52 & 100.00 & - & - & - & - \\
    \midrule
    CorDA  & 64, 256 & 48.15 & 10.97 & 9.35 & 22.82 & 91.93 & 57.09 & 12.86 & 34.98 & 28.25 \\
    LoRA-Null  & 64, 256 & 49.10 & 12.02 & 8.32 & 23.15 & 91.79 & 55.12 & 13.06 & 34.09 & 28.09 \\
    \midrule
    CorDA  & 128, 256 & 46.77 & 10.22 & 9.20 & 22.06 & 89.00 & 58.91 & 14.70 & 36.81 & 28.50 \\
    LoRA-Null  & 128, 256 & 49.03 & 11.52 & 9.01 & 23.19 & 93.00 & 58.76 & 14.06 & 36.41 & 29.06 \\
    \midrule
    CorDA  & 256, 256 & 45.62 & 9.58 & 5.41 & 20.20 & 73.01 & 61.94 & 15.86 & 38.90 & 28.03 \\
    LoRA-Null  & 256, 256 & 46.58 & 10.11 & 8.12 & 21.60 & 84.71 & 62.40 & 16.26 & 39.33 & 29.15 \\
    \bottomrule[1.5pt]
\end{tabular}
\end{subtable}

\caption{LoRA-Null vs. CorDA with varying calibration sizes and ranks on LLaMA-3.2-3B (Math). The ``Avg1'' is the average performance of knowledge preservation. The ``Avg1(Per)'' denotes the average percentage of knowledge preservation. The ``Avg2'' is the average performance of downstream tasks. ``GM'' is the geometric mean of Avg1 and Avg2.}
\label{llama3.2_ab}
\end{table*}

\subsection{Main results on LLaMA-2-7B and LLaMA-3.2-3B}
Tables \ref{llama2_math}, \ref{llama2_code} and \ref{llama2_MT} and \ref{llama3.2_math} present our experimental results on LLaMA-2-7B and LLaMA-3.2-3B across Math, Code and Instruction Following tasks, which demonstrate the superior knowledge preservation of our proposed LoRA-Null, as well as its relatively good performance on downstream tasks. We observe that LoRA-Null achieves the best knowledge preservation performance on Trivia QA and NQ Open compared to the baselines, except in the code for Trivia QA, where LoRA achieves 50.30 and LoRA-Null attains 50.29. For WebQS, LoRA-Null achieves the second-best overall performance. Meanwhile, we can see that LoRA-Null achieves the best average performance in knowledge preservation across the board. In terms of performance relative to the pre-trained models, LoRA-Null achieves an average improvement of $3.35\%$ in knowledge preservation percentage ``Avg(Per)'' compared to CorDA. For downstream tasks, LoRA-Null is second only to PiSSA in overall performance and achieves the best results on Math on LLaMA-2-7B. Although LoRA-Null underperforms CorDA slightly on downstream tasks with LLaMA-3.2-3B, our hyperparameters follow those of CorDA. As shown in Table \ref{llama3.2_ab_sample}, when the sample size of the calibration sets is 1024, LoRA-Null outperforms CorDA in both knowledge preservation and fine-tuning performance. 

\subsection{Deep Discussions of LoRA initialization space}

We further study the space of the LoRA initialization for LoRA-Null, MiLoRA and CorDA. We analyze the components of the down-projection matrices $\mathbf{A}$ on the subspace of $\mathbf{U}$ in Equation \ref{SVD_OF_Xpre} ($\mathbf{U}$ is the left singular matrix of $\mathbf{X}_{\text{pre}}$), as shown in Figure \ref{fig:llama3.2_intro2}. We define the projection matrix as:

\begin{equation}
    \mathbf{P} = \mathbf{A} \mathbf{U} \in \mathbb{R}^{ r \times d_{\text{in}}}.
\end{equation} 

The absolute value sum over columns is computed for each row of $\mathbf{P}$, resulting in a projection absolute value sum vector $\mathbf{pa} \in \mathbb{R}^{d_{\text{in}}}$, where the $i$-th element is given by:
\begin{equation}
    \mathbf{pa}_i = \sum_{j=1}^{r} \left| \mathbf{P}_{j,i} \right|, \quad i = 1, 2, \dots, d_{\text{in}}.
\end{equation}
The horizontal axis is $ \mathbf{U}_{[:,i]} $, while the vertical axis  is $ \mathbf{pa}_i $. A higher value of $ \mathbf{pa}_i $ indicates a larger component of the down-projection matrix $ \mathbf{A} $ onto the subspace of $ \mathbf{U}_{[:,i]}$.

We find that for both CorDA and MiLoRA, their $\mathbf{A}$ do not only lie in $\mathbf{U}_{\text{null}} = \mathbf{U}_{[:,-r:]}$, while the matrix $\mathbf{A}$ of LoRA-Null only lie in $\mathbf{U}_{\text{null}}$. Moreover, the $\mathbf{A}$ in CorDA focuses more on the minor singular vectors of $\mathbf{U}$ compared to MiLoRA. Meanwhile, for CorDA, from a mathematical perspective, $\mathbf{W}_0 \mathbf{X}_{\text{pre}} \mathbf{X}_{\text{pre}}^\top = \mathbf{W}_0 \mathbf{U} \Sigma^2 \mathbf{U}^\top$. This amplifies the components of $\mathbf{W}_0$ along the subspaces in $\mathbf{U}$ with singular values greater than $1$, while attenuating those with singular values less than $1$. As a result, the subspace corresponding to the smallest singular values in the SVD of $\mathbf{W}_0 \mathbf{X}_{\text{pre}} \mathbf{X}_{\text{pre}}^\top$ is dominated by the components of $\mathbf{U}$ associated with small singular values. Meanwhile, $\mathbf{C}^{-1} = \mathbf{U} \Sigma^{-2} \mathbf{U}^\top$ does not change the subspace but amplifies the components corresponding to small singular values. This explains why CorDA outperforms MiLoRA from the perspective of space of LoRA initialization. These further confirm that considering the null space of $\mathbf{X}_{\text{pre}}$ is more effective than that of $\mathbf{W}_0$.

\subsection{Hyperparameter Analysis}

In Table~\ref{llama3.2_ab_sample}, we conduct experiments by varying the number of calibration sets. First, LoRA-Null exhibits greater stability than CorDA, as indicated by its smaller performance range. Second, when using a small calibration set, CorDA suffers from a significant decline in its ability to preserve pre-trained knowledge, whereas LoRA-Null still has strong capability. Third, compared to CorDA, LoRA-Null has significantly better average performance in preserving the pre-trained knowledge and improving downstream task performance across different numbers of calibration sets.

In Table~\ref{llama3.2_ab_rank}, we conduct experiments by varying the rank of LoRA. First, LoRA-Null has a better ability to preserve the pre-trained knowledge than CorDA under different ranks. Second, as the rank increases, CorDA's ability to preserve pre-trained knowledge degrades more rapidly compared to LoRA-Null. Third, as the rank increases, LoRA-Null does not only performs better than CorDA in preserving knowledge but also achieves superior performance on downstream tasks.

\section{Conclusion}
In this paper, we investigate how LoRA initialization preserves pre-trained knowledge. We find that it is crucial for knowledge preservation to make LoRA initialization orthogonal to the pre-trained knowledge, rather than making the residual weights close to the pre-trained weights. We find that
the effective ranks of input activations are much smaller than
those of pre-trained weights. We propose LoRA-Null, a low-rank adaptation method initialized in the null space of the activations of the pre-trained knowledge, which helps preserve LLM knowledge during fine-tuning.  Extensive experiments show that LoRA-Null can achieve good downstream performance and effective knowledge preservation. We hope our research will provide useful insights for future studies on LoRA initialization.

\newpage
\newpage
\clearpage

\section{Acknowledgments}
This research was supported by National Key Research and Development Program of China(NO. 2024YFE0203200), National Natural Science Foundation of China (No. 62476277), CCF-ALIMAMA TECH Kangaroo Fund (No. CCF-ALIMAMA OF 2024008), and Huawei-Renmin University joint program on Information Retrieval. We also acknowledge the support provided by the fund for building worldclass universities (disciplines) of Renmin University of China and by the funds from Beijing Key Laboratory of Big Data Management and Analysis Methods, Gaoling School of Artificial Intelligence, Renmin University of China, from Engineering Research Center of Next-Generation Intelligent Search and Recommendation, Ministry of Education, from Intelligent Social Governance Interdisciplinary Platform, Major Innovation \& Planning Interdisciplinary Platform for the “DoubleFirst Class” Initiative, Renmin University of China, from Public Policy and Decision-making Research Lab of Renmin University of China, and from Public Computing Cloud, Renmin University of China.

\bigskip
%\noindent Thank you for reading these instructions carefully. We look forward to receiving your electronic files!

\bibliography{aaai2026}

@inproceedings{llamapro,
    title = "{LL}a{MA} Pro: Progressive {LL}a{MA} with Block Expansion",
    author = "Wu, Chengyue  and
      Gan, Yukang  and
      Ge, Yixiao  and
      Lu, Zeyu  and
      Wang, Jiahao  and
      Feng, Ye  and
      Shan, Ying  and
      Luo, Ping",
    editor = "Ku, Lun-Wei  and
      Martins, Andre  and
      Srikumar, Vivek",
    booktitle = "Proceedings of the 62nd Annual Meeting of the Association for Computational Linguistics (Volume 1: Long Papers)",
    month = aug,
    year = "2024",
    address = "Bangkok, Thailand",
    publisher = "Association for Computational Linguistics",
    url = "https://aclanthology.org/2024.acl-long.352/",
    doi = "10.18653/v1/2024.acl-long.352",
    pages = "6518--6537"
}

@inproceedings{li-liang-2021-prefix,
    title = "Prefix-Tuning: Optimizing Continuous Prompts for Generation",
    author = "Li, Xiang Lisa  and
      Liang, Percy",
    editor = "Zong, Chengqing  and
      Xia, Fei  and
      Li, Wenjie  and
      Navigli, Roberto",
    booktitle = "Proceedings of the 59th Annual Meeting of the Association for Computational Linguistics and the 11th International Joint Conference on Natural Language Processing (Volume 1: Long Papers)",
    month = aug,
    year = "2021",
    address = "Online",
    publisher = "Association for Computational Linguistics",
    url = "https://aclanthology.org/2021.acl-long.353/",
    doi = "10.18653/v1/2021.acl-long.353",
    pages = "4582--4597"
}

@inproceedings{milora,
    title = "{M}i{L}o{RA}: Harnessing Minor Singular Components for Parameter-Efficient {LLM} Finetuning",
    author = "Wang, Hanqing  and
      Li, Yixia  and
      Wang, Shuo  and
      Chen, Guanhua  and
      Chen, Yun",
    editor = "Chiruzzo, Luis  and
      Ritter, Alan  and
      Wang, Lu",
    booktitle = "Proceedings of the 2025 Conference of the Nations of the Americas Chapter of the Association for Computational Linguistics: Human Language Technologies (Volume 1: Long Papers)",
    month = apr,
    year = "2025",
    address = "Albuquerque, New Mexico",
    publisher = "Association for Computational Linguistics",
    url = "https://aclanthology.org/2025.naacl-long.248/",
    doi = "10.18653/v1/2025.naacl-long.248",
    pages = "4823--4836",
    ISBN = "979-8-89176-189-6"
}

@inproceedings{adapter,
  title = 	 {Parameter-Efficient Transfer Learning for {NLP}},
  author =       {Houlsby, Neil and Giurgiu, Andrei and Jastrzebski, Stanislaw and Morrone, Bruna and De Laroussilhe, Quentin and Gesmundo, Andrea and Attariyan, Mona and Gelly, Sylvain},
  booktitle = 	 {Proceedings of the 36th International Conference on Machine Learning},
  pages = 	 {2790--2799},
  year = 	 {2019},
  editor = 	 {Chaudhuri, Kamalika and Salakhutdinov, Ruslan},
  volume = 	 {97},
  series = 	 {Proceedings of Machine Learning Research},
  month = 	 {09--15 Jun},
  publisher =    {PMLR},
  url = 	 {https://proceedings.mlr.press/v97/houlsby19a.html}
}

@inproceedings{prompt_tuning,
    title = "The Power of Scale for Parameter-Efficient Prompt Tuning",
    author = "Lester, Brian  and
      Al-Rfou, Rami  and
      Constant, Noah",
    editor = "Moens, Marie-Francine  and
      Huang, Xuanjing  and
      Specia, Lucia  and
      Yih, Scott Wen-tau",
    booktitle = "Proceedings of the 2021 Conference on Empirical Methods in Natural Language Processing",
    month = nov,
    year = "2021",
    address = "Online and Punta Cana, Dominican Republic",
    publisher = "Association for Computational Linguistics",
    url = "https://aclanthology.org/2021.emnlp-main.243/",
    doi = "10.18653/v1/2021.emnlp-main.243",
    pages = "3045--3059"
}

@inproceedings{loramoe,
    title = "{L}o{RAM}o{E}: Alleviating World Knowledge Forgetting in Large Language Models via {M}o{E}-Style Plugin",
    author = "Dou, Shihan  and
      Zhou, Enyu  and
      Liu, Yan  and
      Gao, Songyang  and
      Shen, Wei  and
      Xiong, Limao  and
      Zhou, Yuhao  and
      Wang, Xiao  and
      Xi, Zhiheng  and
      Fan, Xiaoran  and
      Pu, Shiliang  and
      Zhu, Jiang  and
      Zheng, Rui  and
      Gui, Tao  and
      Zhang, Qi  and
      Huang, Xuanjing",
    editor = "Ku, Lun-Wei  and
      Martins, Andre  and
      Srikumar, Vivek",
    booktitle = "Proceedings of the 62nd Annual Meeting of the Association for Computational Linguistics (Volume 1: Long Papers)",
    month = aug,
    year = "2024",
    address = "Bangkok, Thailand",
    publisher = "Association for Computational Linguistics",
    url = "https://aclanthology.org/2024.acl-long.106/",
    doi = "10.18653/v1/2024.acl-long.106",
    pages = "1932--1945"
}

@inproceedings{yu2024metamath,
title={MetaMath: Bootstrap Your Own Mathematical Questions for Large Language Models},
author={Longhui Yu and Weisen Jiang and Han Shi and Jincheng YU and Zhengying Liu and Yu Zhang and James Kwok and Zhenguo Li and Adrian Weller and Weiyang Liu},
booktitle={The Twelfth International Conference on Learning Representations},
year={2024},
url={https://openreview.net/forum?id=N8N0hgNDRt}
}

@misc{gsm8k,
      title={Training Verifiers to Solve Math Word Problems}, 
      author={Karl Cobbe and Vineet Kosaraju and Mohammad Bavarian and Mark Chen and Heewoo Jun and Lukasz Kaiser and Matthias Plappert and Jerry Tworek and Jacob Hilton and Reiichiro Nakano and Christopher Hesse and John Schulman},
      year={2021},
      eprint={2110.14168},
      archivePrefix={arXiv},
      primaryClass={cs.LG},
      url={https://arxiv.org/abs/2110.14168}, 
}

@misc{math,
      title={Measuring Mathematical Problem Solving With the MATH Dataset}, 
      author={Dan Hendrycks and Collin Burns and Saurav Kadavath and Akul Arora and Steven Basart and Eric Tang and Dawn Song and Jacob Steinhardt},
      year={2021},
      eprint={2103.03874},
      archivePrefix={arXiv},
      primaryClass={cs.LG},
      url={https://arxiv.org/abs/2103.03874}, 
}

@inproceedings{meng2024pissa,
title={Pi{SSA}: Principal Singular Values and Singular Vectors Adaptation of Large Language Models},
author={Fanxu Meng and Zhaohui Wang and Muhan Zhang},
booktitle={The Thirty-eighth Annual Conference on Neural Information Processing Systems},
year={2024},
url={https://openreview.net/forum?id=6ZBHIEtdP4}
}

@inproceedings{hu2022lora,
    title={Lo{RA}: Low-Rank Adaptation of Large Language Models},
    author={Edward J Hu and yelong shen and Phillip Wallis and Zeyuan Allen-Zhu and Yuanzhi Li and Shean Wang and Lu Wang and Weizhu Chen},
    booktitle={International Conference on Learning Representations},
    year={2022},
    url={https://openreview.net/forum?id=nZeVKeeFYf9}
}

@inproceedings{
yang2024corda,
title={Cor{DA}: Context-Oriented Decomposition Adaptation of Large Language Models for Task-Aware Parameter-Efficient Fine-tuning},
author={Yibo Yang and Xiaojie Li and Zhongzhu Zhou and Shuaiwen Leon Song and Jianlong Wu and Liqiang Nie and Bernard Ghanem},
booktitle={The Thirty-eighth Annual Conference on Neural Information Processing Systems},
year={2024},
url={https://openreview.net/forum?id=Gi00NVru6n}
}

@article{loralearnless,
title={Lo{RA} Learns Less and Forgets Less},
author={Dan Biderman and Jacob Portes and Jose Javier Gonzalez Ortiz and Mansheej Paul and Philip Greengard and Connor Jennings and Daniel King and Sam Havens and Vitaliy Chiley and Jonathan Frankle and Cody Blakeney and John Patrick Cunningham},
journal={Transactions on Machine Learning Research},
issn={2835-8856},
year={2024},
url={https://openreview.net/forum?id=aloEru2qCG},
note={Featured Certification}
}

@inproceedings{triviaqa,
    title = "{T}rivia{QA}: A Large Scale Distantly Supervised Challenge Dataset for Reading Comprehension",
    author = "Joshi, Mandar  and
      Choi, Eunsol  and
      Weld, Daniel  and
      Zettlemoyer, Luke",
    editor = "Barzilay, Regina  and
      Kan, Min-Yen",
    booktitle = "Proceedings of the 55th Annual Meeting of the Association for Computational Linguistics (Volume 1: Long Papers)",
    month = jul,
    year = "2017",
    address = "Vancouver, Canada",
    publisher = "Association for Computational Linguistics",
    url = "https://aclanthology.org/P17-1147/",
    doi = "10.18653/v1/P17-1147",
    pages = "1601--1611"
}

@inproceedings{nq_open,
    title = "Latent Retrieval for Weakly Supervised Open Domain Question Answering",
    author = "Lee, Kenton  and
      Chang, Ming-Wei  and
      Toutanova, Kristina",
    editor = "Korhonen, Anna  and
      Traum, David  and
      M{\`a}rquez, Llu{\'i}s",
    booktitle = "Proceedings of the 57th Annual Meeting of the Association for Computational Linguistics",
    month = jul,
    year = "2019",
    address = "Florence, Italy",
    publisher = "Association for Computational Linguistics",
    url = "https://aclanthology.org/P19-1612/",
    doi = "10.18653/v1/P19-1612",
    pages = "6086--6096"
}

@inproceedings{webqs,
    title = "Semantic Parsing on {F}reebase from Question-Answer Pairs",
    author = "Berant, Jonathan  and
      Chou, Andrew  and
      Frostig, Roy  and
      Liang, Percy",
    editor = "Yarowsky, David  and
      Baldwin, Timothy  and
      Korhonen, Anna  and
      Livescu, Karen  and
      Bethard, Steven",
    booktitle = "Proceedings of the 2013 Conference on Empirical Methods in Natural Language Processing",
    month = oct,
    year = "2013",
    address = "Seattle, Washington, USA",
    publisher = "Association for Computational Linguistics",
    url = "https://aclanthology.org/D13-1160/",
    pages = "1533--1544"
}

@misc{cobbe2021training,
  title={Training verifiers to solve math word problems, 2021},
  author={Cobbe, Karl and Kosaraju, Vineet and Bavarian, Mohammad and Chen, Mark and Jun, Heewoo and Kaiser, Lukasz and Plappert, Matthias and Tworek, Jerry and Hilton, Jacob and Nakano, Reiichiro and others},
  journal={URL https://arxiv. org/abs/2110.14168},
  year={2021}
}

@misc{humaneval,
  title={Evaluating large language models trained on code},
  author={Chen, Mark and Tworek, Jerry and Jun, Heewoo and Yuan, Qiming and Pinto, Henrique Ponde De Oliveira and Kaplan, Jared and Edwards, Harri and Burda, Yuri and Joseph, Nicholas and Brockman, Greg and others},
  journal={arXiv preprint arXiv:2107.03374},
  year={2021}
}

@misc{mbpp,
      title={Program Synthesis with Large Language Models}, 
      author={Jacob Austin and Augustus Odena and Maxwell Nye and Maarten Bosma and Henryk Michalewski and David Dohan and Ellen Jiang and Carrie Cai and Michael Terry and Quoc Le and Charles Sutton},
      year={2021},
      eprint={2108.07732},
      archivePrefix={arXiv},
      primaryClass={cs.PL},
      url={https://arxiv.org/abs/2108.07732}, 
}

@inproceedings{mtbench,
title={Judging {LLM}-as-a-Judge with {MT}-Bench and Chatbot Arena},
author={Lianmin Zheng and Wei-Lin Chiang and Ying Sheng and Siyuan Zhuang and Zhanghao Wu and Yonghao Zhuang and Zi Lin and Zhuohan Li and Dacheng Li and Eric Xing and Hao Zhang and Joseph E. Gonzalez and Ion Stoica},
booktitle={Thirty-seventh Conference on Neural Information Processing Systems Datasets and Benchmarks Track},
year={2023},
url={https://openreview.net/forum?id=uccHPGDlao}
}

@inproceedings{
xu2024wizardlm,
title={Wizard{LM}: Empowering Large Pre-Trained Language Models to Follow Complex Instructions},
author={Can Xu and Qingfeng Sun and Kai Zheng and Xiubo Geng and Pu Zhao and Jiazhan Feng and Chongyang Tao and Qingwei Lin and Daxin Jiang},
booktitle={The Twelfth International Conference on Learning Representations},
year={2024},
url={https://openreview.net/forum?id=CfXh93NDgH}
}

@book{strang2022introduction,
  title={Introduction to linear algebra},
  author={Strang, Gilbert},
  year={2022},
  publisher={SIAM}
}

@inproceedings{pmlr-v235-wei24h,
  title = 	 {Magicoder: Empowering Code Generation with {OSS}-Instruct},
  author =       {Wei, Yuxiang and Wang, Zhe and Liu, Jiawei and Ding, Yifeng and Zhang, Lingming},
  booktitle = 	 {Proceedings of the 41st International Conference on Machine Learning},
  pages = 	 {52632--52657},
  year = 	 {2024},
  editor = 	 {Salakhutdinov, Ruslan and Kolter, Zico and Heller, Katherine and Weller, Adrian and Oliver, Nuria and Scarlett, Jonathan and Berkenkamp, Felix},
  volume = 	 {235},
  series = 	 {Proceedings of Machine Learning Research},
  month = 	 {21--27 Jul},
  publisher =    {PMLR},
  url = 	 {https://proceedings.mlr.press/v235/wei24h.html}
}

@inproceedings{
loshchilov2018decoupled,
title={Decoupled Weight Decay Regularization},
author={Ilya Loshchilov and Frank Hutter},
booktitle={International Conference on Learning Representations},
year={2019},
url={https://openreview.net/forum?id=Bkg6RiCqY7},
}

@inproceedings{lin2024awq,
 author = {Lin, Ji and Tang, Jiaming and Tang, Haotian and Yang, Shang and Chen, Wei-Ming and Wang, Wei-Chen and Xiao, Guangxuan and Dang, Xingyu and Gan, Chuang and Han, Song},
 booktitle = {Proceedings of Machine Learning and Systems},
 editor = {P. Gibbons and G. Pekhimenko and C. De Sa},
 pages = {87--100},
 title = {AWQ: Activation-aware Weight Quantization for On-Device LLM Compression and Acceleration},
 url = {https://proceedings.mlsys.org/paper_files/paper/2024/file/42a452cbafa9dd64e9ba4aa95cc1ef21-Paper-Conference.pdf},
 volume = {6},
 year = {2024}
}

@inproceedings{
wang2025svdllm,
title={{SVD}-{LLM}: Truncation-aware Singular Value Decomposition for Large Language Model Compression},
author={Xin Wang and Yu Zheng and Zhongwei Wan and Mi Zhang},
booktitle={The Thirteenth International Conference on Learning Representations},
year={2025},
url={https://openreview.net/forum?id=LNYIUouhdt}
}

@inproceedings{erank,
  author={Roy, Olivier and Vetterli, Martin},
  booktitle={2007 15th European Signal Processing Conference}, 
  title={The effective rank: A measure of effective dimensionality}, 
  year={2007},
  volume={},
  number={},
  pages={606-610},
  doi={}}

@inproceedings{
tang2025adept,
title={{AD}e{PT}: Adaptive Decomposed Prompt Tuning for Parameter-Efficient Fine-tuning},
author={Pengwei Tang and Xiaolin Hu and Yong Liu},
booktitle={The Thirteenth International Conference on Learning Representations},
year={2025},
url={https://openreview.net/forum?id=fswihJIYbd}
}

@inproceedings{ijcai2025p760,
  title     = {Theoretical Insights into Fine-Tuning Attention Mechanism: Generalization and Optimization},
  author    = {Yao, Xinhao and Qian, Hongjin and Hu, Xiaolin and Xu, Gengze and Liu, Wei and Luan, Jian and Wang, Bin and Liu, Yong},
  booktitle = {Proceedings of the Thirty-Fourth International Joint Conference on
               Artificial Intelligence, {IJCAI-25}},
  publisher = {International Joint Conferences on Artificial Intelligence Organization},
  editor    = {James Kwok},
  pages     = {6830--6838},
  year      = {2025},
  month     = {8},
  note      = {Main Track},
  doi       = {10.24963/ijcai.2025/760},
  url       = {https://doi.org/10.24963/ijcai.2025/760},
}

\clearpage

\section{The proof of Theorem \ref{theorem1}}

We analyze the following optimization formula:
\begin{align}
\begin{aligned}
    & \min_{\mathbf{W}_0'}  \|\mathbf{W}_0'  - \mathbf{W}_0\|_{\mathrm{F}}, \\ 
    &\text{s.t.} \quad \text{rank}\left(\mathbf{W}_0\right)=R, \quad \text{rank}\left(\mathbf{W}_0'\right)= R-r. 
    \label{optimation1}
\end{aligned}
\end{align} 

If $ \verb|SVD|(\mathbf{W}_0) = \mathbf{\hat{U}} \mathbf{\hat{\Sigma}} \mathbf{\hat{V}}^\top$, we can get the solution
\begin{equation}
    \mathbf{W}'_0 = \mathbf{\hat{U}}_{[:,:R-r]}\mathbf{\hat{\Sigma}}_{[:R-r]} \mathbf{\hat{V}^\top}_{[:R-r,:]}. \label{sol:1}
\end{equation}

\textbf{Equation \ref{sol:1} is aligned with the MiLoRA.}

\section{The proof of Theorem \ref{theorem2}}

We analyze the following optimization formula:
\begin{align}
\begin{aligned}
    & \min_{\mathbf{W}_0'}  \|\mathbf{W}_0' \mathbf{X}_{\textbf{pre}} - \mathbf{W}_0 \mathbf{X}_{\textbf{pre}}\|_{\mathrm{F}}, \\ &\text{s.t.} \quad \text{rank}\left(\mathbf{W}_0\right)=R, \quad \text{rank}\left(\mathbf{W'}_0\right)= R-r. \label{optimation2}
\end{aligned}
\end{align}
 To find the optimal $\mathbf{W'}_0$, we expand the Frobenius norm:

\begin{align}
\begin{aligned}  
& \|\mathbf{W}_0'\mathbf{X}_{\text{pre}} - \mathbf{W}_0\mathbf{X}_{\text{pre}}\|_F^2 \\ &= \text{tr}\left((\mathbf{W}_0'\mathbf{X}_{\text{pre}} - \mathbf{W}_0\mathbf{X}_{\text{pre}})^\top (\mathbf{W}_0'\mathbf{X}_{\text{pre}} - \mathbf{W}_0\mathbf{X}_{\text{pre}})\right),
\end{aligned}
\end{align}

which $\text{tr}$ means the trace of the matrix. The trace of the matrix $\mathbf{G} = [g_{ij}]$ is defined as
\begin{equation}
    \text{tr}{\mathbf{\left(G\right)}} = \sum_{i} g_{ii}.
\end{equation}

To get the optimal $\mathbf{W}_0'$, we can differentiate $\text{tr}\left((\mathbf{W}'_0\mathbf{X}_{\text{pre}} - \mathbf{W}_0\mathbf{X}_{\text{pre}})^\top (\mathbf{W}'_0\mathbf{X}_{\text{pre}} - \mathbf{W}_0\mathbf{X}_{\text{pre}})\right)$ with respect to $\mathbf{W}_0'$, which is formulated as

\begin{align} 
\begin{aligned}
    & \frac{\partial}{\partial \mathbf{W}_0'} \left[ \text{tr} \left( \left( \mathbf{W}_0'\mathbf{X}_{\text{pre}} - \mathbf{W}_0\mathbf{X}_{\text{pre}} \right)^\top \left( \mathbf{W}_0'\mathbf{X}_{\text{pre}} - \mathbf{W}_0\mathbf{X}_{\text{pre}} \right) \right) \right] 
    \\ &= 2 \left( \mathbf{W}_0'\mathbf{X}_{\text{pre}} - \mathbf{W}_0\mathbf{X}_{\text{pre}} \right) \mathbf{X}_{\text{pre}}^\top.
\end{aligned}
\end{align}

From the condition that the derivative of the optimal $\mathbf{W}'$ equals the zero vector, we obtain:
\begin{equation}
    2 \left( \mathbf{W}'\mathbf{X}_{\text{pre}} - \mathbf{W}\mathbf{X}_{\text{pre}} \right) \mathbf{X}_{\text{pre}}^\top = \mathbf{0}.
\end{equation}

We have

\begin{equation}
    \mathbf{W}_0' \mathbf{X}_{\text{pre}}\mathbf{X}_{\text{pre}}^\top = \mathbf{W}_0 \mathbf{X}_{\text{pre}}\mathbf{X}_{\text{pre}}^\top.
\end{equation}

%If $\left(\mathbf{X}_{\text{pre}}\mathbf{X}_{\text{pre}}^\top\right)$ is invertible, we have
We have
\begin{equation}
    \mathbf{W}_0' = \mathbf{W}_0 \mathbf{X}_{\text{pre}}\mathbf{X}_{\text{pre}}^\top \left(\mathbf{X}_{\text{pre}}\mathbf{X}_{\text{pre}}^\top\right)^{+}, \label{ref:solution2}
\end{equation}
where $^{+}$ denotes the pseudo-inverse of a matrix.

To meet the low-rank constraint, we can use SVD to approximate the solution. We can first use SVD for $\mathbf{W}_0$, $\mathbf{W}_0\mathbf{X}_{\text{pre}}$ or $\mathbf{W}_0 \mathbf{X}_{\text{pre}} \mathbf{X}_{\text{pre}}^\top$, then multiply this approximation by $\mathbf{X}_{\text{pre}}\mathbf{X}_{\text{pre}}^\top \left(\mathbf{X}_{\text{pre}}\mathbf{X}_{\text{pre}}^\top\right)^{+}$, $\mathbf{X}_{\text{pre}}^\top \left(\mathbf{X}_{\text{pre}}\mathbf{X}_{\text{pre}}^\top\right)^{+}$ or $\left(\mathbf{X}_{\text{pre}}\mathbf{X}_{\text{pre}}^\top\right)^{+}$. \textbf{Making SVD for $\mathbf{W}_0\mathbf{X}_{\text{pre}}\mathbf{X}_{\text{pre}}^\top$ and multiplying the approximation by $\left(\mathbf{X}_{\text{pre}}\mathbf{X}_{\text{pre}}^\top\right)^{+}$ is aligned with CorDA.}

We complete the proof.

\section{The proof of Theorem \ref{theorem:3}}

First, it is clear that LoRA-Null is not the solution of Equation~\ref{eq:formulation1}, because Equation~\ref{eq:formulation1} does not contain $\mathbf{X}_{\text{pre}}$.

Second, in fact, for Equation \ref{optimation2}, using SVD for $\mathbf{W}_0$ is the same as MiLoRA. It means that MiLoRA can also be seen as a solution of Equation \ref{optimation2}. The choice of $\mathbf{W}_0\mathbf{X}_{\text{pre}}\mathbf{X}_{\text{pre}}^\top$ is not theoretically guaranteed to provide the optimal approximation. However, CorDA use experiments to demonstrate using $\mathbf{W}_0\mathbf{X}_{\text{pre}}\mathbf{X}_{\text{pre}}^\top$ are better than using $\mathbf{W}_0$. Whether using $\mathbf{W}_0\mathbf{X}_{\text{pre}}$ is better remains an open question and warrants further investigation, but this is not related to the core idea of this paper. \textbf{Also, we can observe that the $\mathbf{W}_0 - \mathbf{W}_0 \mathbf{U}_{\text{null}} \mathbf{U}_{\text{null}}^\top$ is not the solution of Equation~\eqref{eq:formulation2}.}

We complete the proof.

\section{Proof That the Column  of $\mathbf{A}^\top$ Lies in the Column Space of $\mathbf{U}_{\text{null}}$}
\label{proof_A_U_null}
%\textbf{Step 1: Determine the column space of $\mathbf{A}$.}

\begin{tcolorbox}[colback=gray!20,colframe=gray]
\begin{theorem}%[3.1]
Let $\mathbf{W}_0 \in \mathbb{R}^{n \times m}$, $\mathbf{U}_\text{null} \in \mathbb{R}^{m \times r}$, $r < \min(n,m)$ and $\mathbf{U}_\text{null} \mathbf{U}^\top_{\text{null}} = \mathbf{I}_r$. Make SVD for $\verb|SVD|(\mathbf{W}_0\mathbf{U}_\text{null}\mathbf{U}^\top_{\text{null}}) = \mathbf{U}'\mathbf{\Sigma}'\mathbf{V}'$. Let $\mathbf{A} = \sqrt{\mathbf{\Sigma}'_{[:r]}} \left(\mathbf{V}_{[:,:r]}'\right)^\top = \sqrt{\mathbf{\Sigma}'_{[:r]}} \mathbf{V}'^\top _{[:r,:]}$. Then, the column space of $\mathbf{A}^T$ lies in the column space of $\mathbf{U}_{\text{null}}$, \textit{i.e.}, $\text{Col}(\mathbf{A}^\top) = \text{Col}(\mathbf{U}_{\text{null}})$.
\end{theorem}
\end{tcolorbox}
\begin{proof}
\textbf{Step 1: }
From the definition:
\begin{equation}
        \mathbf{A} = \sqrt{\mathbf{\Sigma}'_{[:r]}} \left(\mathbf{V}_{[:,:r]}'\right)^\top = \sqrt{\mathbf{\Sigma}'_{[:r]}} \mathbf{V}'^\top _{[:r,:]}.
\end{equation}

Let $\mathbf{Z} = \left(\sqrt{\mathbf{\Sigma}'_{[:r]}} \mathbf{V}_{[:r,:]}'^\top\right)^\top = \mathbf{V}_{[:,:r]}'\left(\sqrt{\mathbf{\Sigma}'^\top_{[:r]}}\right) $, then $\mathbf{A}^\top = \mathbf{Z}$. Since $\left(\sqrt{\mathbf{\Sigma}'^\top_{[:r]}}\right)$ is a diagonal matrix, it only scales the columns of $\mathbf{V}_{[:,:r]}'$ without altering their span. Therefore, the column space of $\mathbf{A}^\top$ lies in the column space of $\mathbf{V}_{[:,:r]}'$:
\begin{equation}
\text{Col}(\mathbf{A}^\top) \subseteq \text{Col}(\mathbf{V}_{[:,:r]}').
\end{equation}

\textbf{Step 2: }
%We first perform SVD for:
%\begin{equation}
%\verb|SVD|(\mathbf{W}_0)  = \mathbf{U}'' \mathbf{\Sigma}'' (\mathbf{V}'')^\top,
%\end{equation}
%\textbf{Step 2: Analyze the column space of $\mathbf{V}'$.}
%Thus, we can get
%\begin{equation}
%\mathbf{W}_0 \mathbf{U}_{\text{null}} \mathbf{U}_{\text{null}}\top = \mathbf{U}'' \mathbf{\Sigma}'' (\mathbf{V}'')\top \mathbf{U}_{\text{null}} \mathbf{U}_{\text{null}}\top,
%\end{equation}
Let $\mathbf{W}_0 \mathbf{U}_{\text{null}} = \mathbf{\hat{W}}$, and make SVD for $\mathbf{\hat{W}} = \mathbf{P} \mathbf{\Sigma} \mathbf{Q}^\top$. Then, $\mathbf{\hat{W}} U_{\text{null}}^\top = \mathbf{P} \mathbf{\Sigma} \mathbf{Q}^\top \mathbf{U}_{\text{null}}^\top$. Let $\mathbf{L} = \mathbf{U}_{\text{null}} \mathbf{Q} $. Then $\mathbf{L}^\top \mathbf{L}^\top = \mathbf{Q} ^\top\mathbf{U}_{\text{null}} ^\top\mathbf{U}_{\text{null}}\mathbf{Q} = \mathbf{Q}^\top \mathbf{Q} = \mathbf{I}_r$, which means the column vectors of $\mathbf{L}$ are orthogonal. It indicates that $\mathbf{P} \mathbf{\Sigma} \mathbf{L}^\top$ is the SVD for $\mathbf{\hat{W}} \mathbf{U}_{\text{null}}^\top = \mathbf{W}_0 \mathbf{U}_{\text{null}}  \mathbf{U}_{\text{null}}^\top $. That means that $\mathbf{V}' = \mathbf{L} = \mathbf{U}_{\text{null}} \mathbf{Q}$. Thus, $ \mathbf{V}' $ is a linear combination of the column vectors of the matrix $ \mathbf{U}_{\text{null}} $, which is formulated as 
$$
\text{Col}(\mathbf{V}')  \subseteq \text{Col} (\mathbf{U}_{\text{null}}). 
$$

\textbf{Step 3:}
Combining the results from Step 1 and Step 2:
\begin{equation}
\text{Col}(\mathbf{A}^\top) \subseteq \text{Col}\left(\mathbf{V}_{[:,:r]}'\right) \subseteq \text{Col}(\mathbf{V}') \subseteq \text{Col}(\mathbf{U}_{\text{null}}).
\end{equation}

Thus, the column space of $\mathbf{A}^\top$ lies in the column space of $\mathbf{U}_{\text{null}}$.

We complete the proof.
\end{proof}

\section{The corresponding experiments about Theorem~\ref{theorem1}, Theorem~\ref{theorem2} and Theorem~\ref{theorem:3}}

\begin{figure*}[t]
	\centering
	\begin{subfigure}{0.3\textwidth}
		\centering
		% include first image
		\includegraphics[width=1.\linewidth]{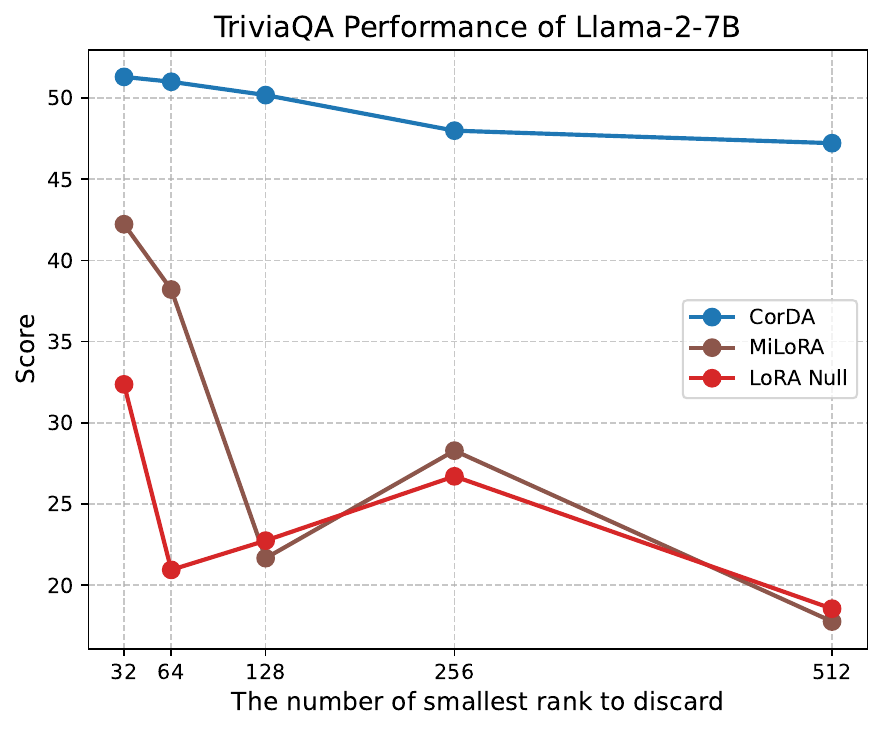}
		\vspace{-2mm}  
		%\caption{Perplexity on Wikitext-2}
		\label{fig:llama2_triviaqa}
	\end{subfigure}
	\begin{subfigure}{0.3\textwidth}
		\centering
		% include second image
		\includegraphics[width=1.\linewidth]{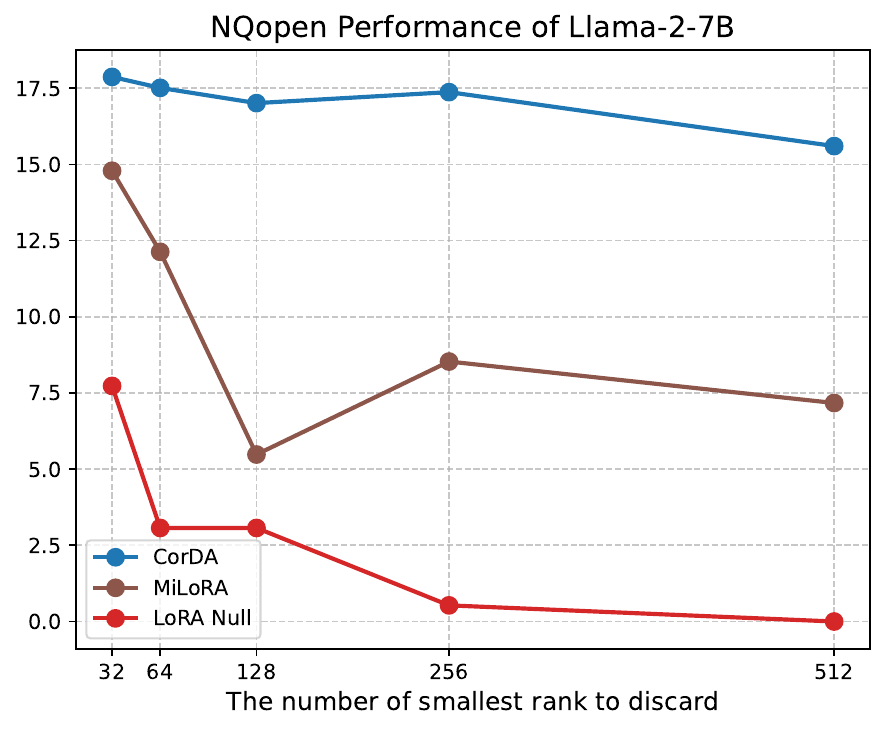}  
		\vspace{-2mm}
		%\caption{Perplexity on PTB}
		\label{fig:llama2_nqopen}
	\end{subfigure}	
    \begin{subfigure}{0.3\textwidth}
		\centering
		% include second image
		\includegraphics[width=1.\linewidth]{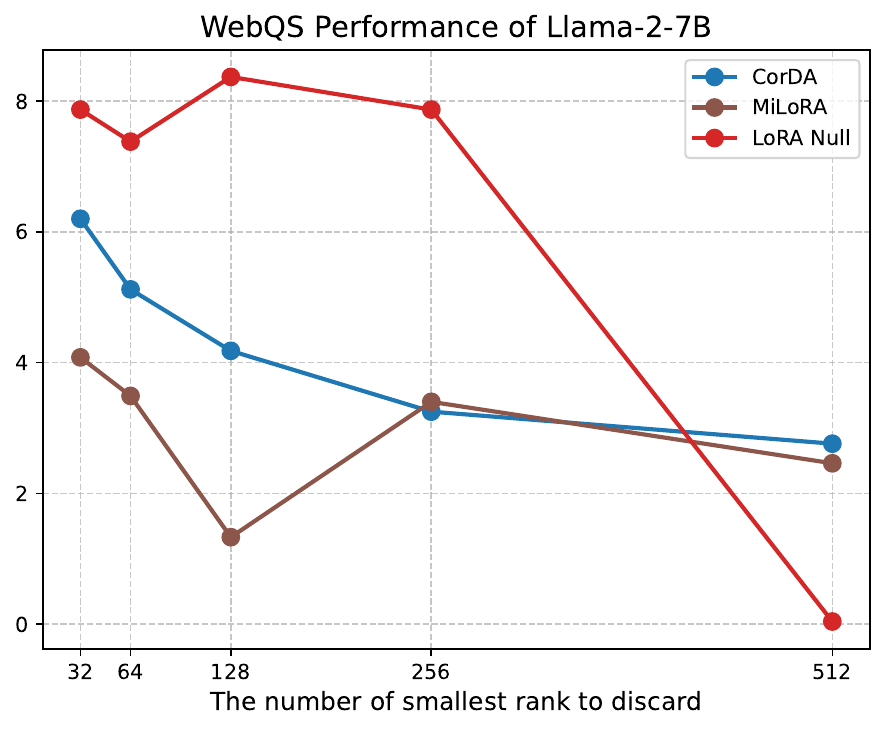}  
		\vspace{-2mm}
		%\caption{Perplexity on PTB}
		\label{fig:llama2_webqs}
	\end{subfigure}	
        %\subcaption{LLaMA-2-7B}
    	\begin{subfigure}{0.3\textwidth}
		\centering
		% include first image
		\includegraphics[width=1.\linewidth]{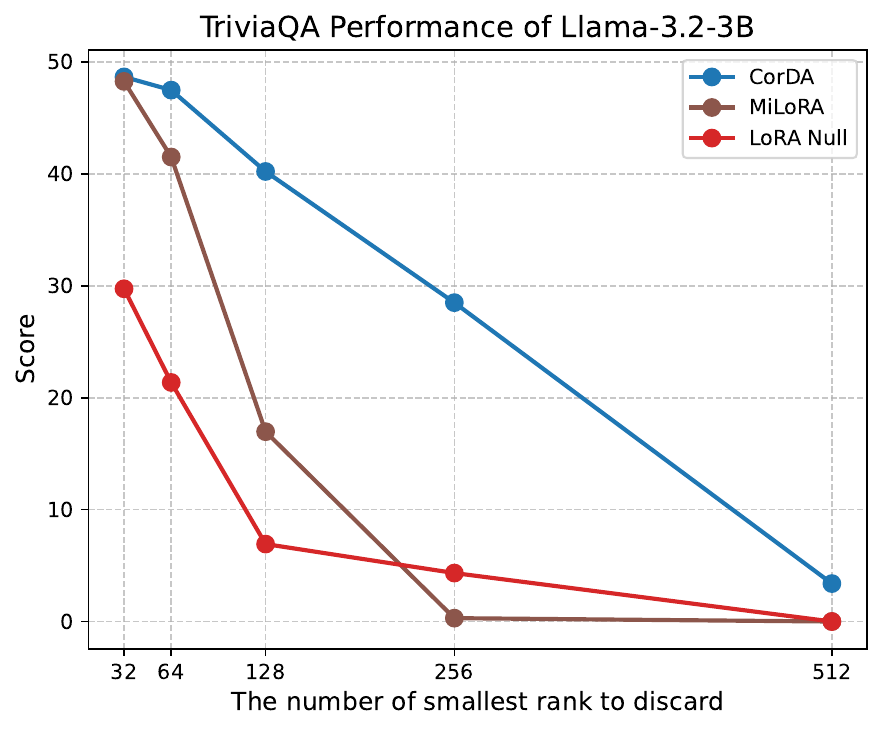}
		\vspace{-2mm}  
		%\caption{Perplexity on Wikitext-2}
		\label{fig:llama3.2_triviaqa}
	\end{subfigure}
	\begin{subfigure}{0.3\textwidth}
		\centering
		% include second image
		\includegraphics[width=1.\linewidth]{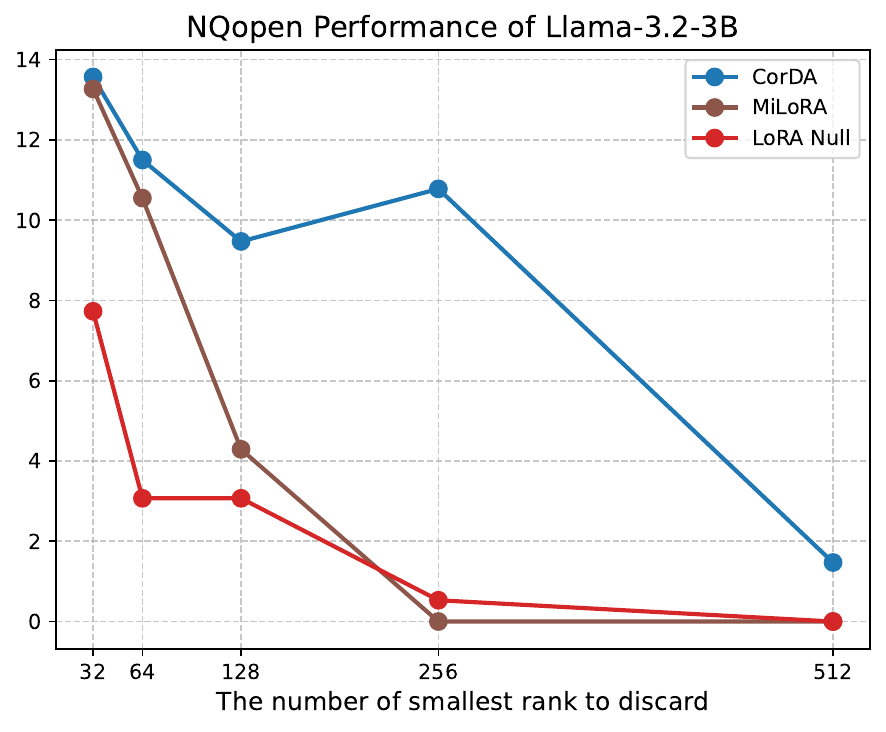}  
		\vspace{-2mm}
		%\caption{Perplexity on PTB}
		\label{fig:llama3.2_nqopen}
	\end{subfigure}	
    \begin{subfigure}{0.3\textwidth}
		\centering
		% include second image
		\includegraphics[width=1.\linewidth]{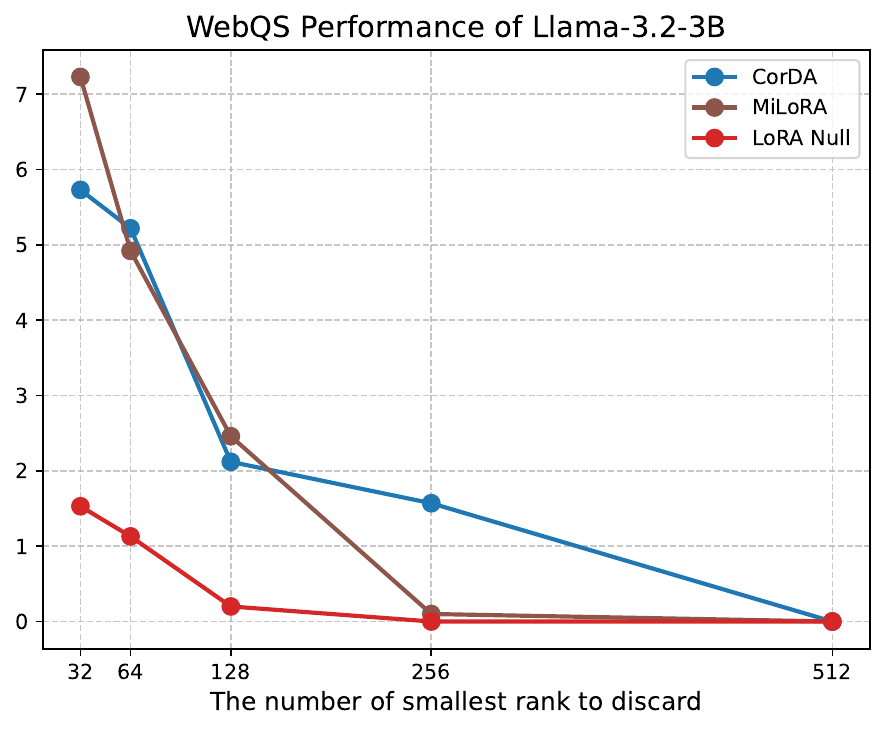}  
		\vspace{-2mm}
		%\caption{Perplexity on PTB}
		\label{fig:llama3.2_webqs}
	\end{subfigure}	\
        %\subcaption{LLaMA-3.2-3B}
        \caption{Score (higher is better) on Trivia QA, NQ open
 and WebQS after decomposing the ``LLaMA-2-7B'' and ``LLaMA-3.2-3B'' weights and reconstruction discarding the LoRA adapters ($r$ is the rank of LoRA adapters). We compare our proposed LoRA-Null with CorDA and MiLoRA.}
	\label{fig:llama_intro1}
	%\vspace{-1mm}
\end{figure*}

\begin{table*}[th!]
\centering
\small % ✅ 显式设为 10pt Roman 字体

% ✅ 用 subtable 包裹，创建子表格环境
\begin{subtable}{\linewidth}
\centering
\setlength{\tabcolsep}{1mm}
\begin{tabular}{l|c|ccccc|ccc|c}
    \toprule[1.5pt]
    Method & \textbf{\#Params} & Trivia QA & NQ open & WebQS & Avg1& Avg1(Per)& GSM8k & Math & Avg2 &GM \\
    \midrule
    Gemma3-1B & - & 32.52 & 9.53 & 6.89 & - & - & - \\
    \midrule
    MiLoRA & 104M & 21.60 & 5.26 & 4.18 & 10.35 & 60.76 & 22.44 & 3.26 & 12.85 & 11.53 \\
    CorDA  & 104M & 21.67 & 5.07 & 4.43 & 10.39 & 61.38 & 22.44 & 3.16 & 12.80 & 11.53 \\
    LoRA-Null ($\Sigma' = \Sigma'$)  & 104M & 29.73 & 6.76 & 5.51& 14.00 & 80.78  & 16.91 & 2.64 & 9.78 & 11.70 \\
    LoRA-Null ($\Sigma' = 2\Sigma'$) & 104M & 27.44 & 6.18 & 5.71 & 13.11 & 77.37 &  19.03 & 2.72 & 10.88 &\underline{11.94} \\
    LoRA-Null ($\Sigma' = 4\Sigma'$) & 104M & 22.73 & 5.12 & 5.12 & 10.99 & 65.98 &  22.90 & 3.16 & 13.03 &\textbf{11.97} \\
    \bottomrule[1.5pt]
\end{tabular}
\end{subtable}

\caption{Performance comparison of LoRA-Null with baseline methods on Gemma-3-1B for Math.}
\label{gemma3_1B_result}
\end{table*}

Only considering the frozen residual weight $\mathbf{W}'_0$, LoRA-Null may not preserve the pre-trained knowledge well. In Figure \ref{fig:llama_intro1}, we evaluate the performance of LoRA-Null, CorDA, and MiLoRA on Trivia QA, NQ Open, and WebQS after reconstructing the weights of LLaMA-2-7B and LLaMA-3.2-3B discarding the LoRA adapters. We can find that LoRA-Null performs worse than MiLoRA, and perform much worse than CorDA. However, in the subsequent experiments as shown in Tables~\ref{llama2_math}, \ref{llama2_code}, \ref{llama2_MT}, and \ref{llama3.2_math}, LoRA-Null demonstrates a better ability to preserve pre-trained knowledge. \textbf{This indicates that it is crucial for knowledge preservation to make LoRA initialization orthogonal to the pre-trained knowledge, rather than making the residual weights close to the pre-trained weights}.

\section{Results on Gemma-3-1B}
Table \ref{gemma3_1B_result} shows the results on Gemma-3-1B. First, we find that the vanilla LoRA-Null can preserve most knowledge while achieving low downstream tasks. Thus, we scale the $\mathbf{\Sigma}'$ to do further study. When the scale of $\mathbf{\Sigma}'$ increases, knowledge preservation worsens while downstream task performance improves. 
When LoRA-Null achieves comparable downstream task performance to CorDA and MiLoRA, it demonstrates stronger knowledge preservation. This indicates that using the null space of activations is key to knowledge preservation.

\begin{table*}[t!]
\centering
\small 

\setlength{\tabcolsep}{1mm} 
\begin{tabular}{l|c|c|c|c|c|c|c}
\toprule
 & qproj & kproj & vproj & oproj & gateproj & upproj & downproj \\
\midrule
$\mathbf{W}_0$ (0) & 1214.11 & 548.30 & 892.36 & 2090.62 & 2828.59 & 2871.72 & 2883.30 \\
\midrule
$\mathbf{X}_{\text{pre}}$ (0) & 101.28 & 101.28 & 101.28 & 75.60 & 553.59 & 553.59 & 758.06 \\
\midrule
$\mathbf{W}_0$ (1) & 1788.42 & 714.95 & 941.63 & 2215.13 & 2808.90 & 2858.23 & 2882.98 \\
\midrule
$\mathbf{X}_{\text{pre}}$ (1) & 85.63 & 85.63 & 85.63 & 227.29 & 657.79 & 657.79 & 1.27 \\
\midrule
$\mathbf{W}_0$ (27) & 2089.48 & 851.92 & 943.69 & 2265.90 & 2748.93 & 2757.13 & 2838.64 \\
\midrule
$\mathbf{X}_{\text{pre}}$ (27) & 194.22 & 194.22 & 194.22 & 72.82 & 302.78 & 302.78 & 79.97 \\
\bottomrule
\end{tabular}

\caption{The effective rank of $\mathbf{X}_{\text{pre}}$ and $\mathbf{W}_0$ on LLaMA-3.2-3B. The number in ``()'' indicates the layer index.}
\label{tab:effecitve_ranks_complete}
\end{table*}

\begin{table*}[t!]
\centering
\small % ✅ 显式设为 10pt Roman 字体

% --- Subtable (a): Different number of calibration sets ---
\begin{subtable}{\linewidth}
\centering
\subcaption{Different number of calibration sets}
\label{llama3.2_ab_sample_appendix}
\setlength{\tabcolsep}{1mm} % ✅ 适当减小列间距替代缩放
\begin{tabular}{l|c|ccccc|ccc|c}
    \toprule[1.5pt]
    Method & \textbf{\#Rank Size, \#Samples Size} & Trivia QA & NQ open & WebQS & Avg1 & Avg1(Per) & GSM8k & Math & Avg2 & GM \\
    \midrule
    LLaMA-3.2-3B & - & 50.77 & 13.55 & 9.25 & 24.52 & 100.00 & - & - & - & - \\
    \midrule
    CorDA  & 128, 64 & 42.72 & 10.00 & 4.53 & 19.08 & 68.97 & 62.32 & 15.40 & 38.86 & 27.23 \\
    LoRA-Null  & 128, 64 & 48.69 & 12.33 & 9.06 & 23.36 & 94.95 & 59.59 & 13.94 & 36.77 & 29.30 \\
    \midrule
    CorDA  & 128, 256 & 46.77 & 10.22 & 9.20 & 22.06 & 89.00 & 58.91 & 14.70 & 36.81 & 28.51 \\
    LoRA-Null  & 128, 256 & 49.03 & 11.52 & 9.01 & 23.19 & 93.00 & 58.76 & 14.06 & 36.41 & 29.07 \\
    \midrule
    CorDA  & 128, 512 &  48.48	& 10.66 & 	9.79	& 22.98 & 91.39 & 58.76	& 14.78	& 36.77 &  29.07 \\
    LoRA-Null  & 128, 512 &  48.45	& 11.05	& 8.81	& 22.77 & 90.74 & 59.14	& 14.52	& 36.83 & 28.96 \\
    \midrule
    CorDA  & 128, 768 &  48.17 & 11.50 & 8.56 & 22.74 &  90.76 &57.46 & 14.26 & 35.86 & 28.56\\
    LoRA-Null  & 128, 768 & 48.79 & 11.63 & 9.15 & 23.19 & 93.62 & 58.76 & 14.3 &  36.53 &  29.10 \\
    \midrule
    CorDA  & 128, 1024 & 48.46 & 9.89 & 9.30 & 22.55 & 89.48 & 58.15 & 14.50 & 36.33 & 28.61 \\
    LoRA-Null  & 128, 1024 & 48.98 & 11.14 & 10.78 & 23.63 & 92.90 & 59.59 & 14.54 & 37.07 & 29.59 \\
    \bottomrule[1.5pt]
\end{tabular}
\end{subtable}
\caption{LoRA-Null vs. CorDA with varying calibration sizes on LLaMA-3.2-3B (Math). The ``Avg1'' is the average performance of knowledge preservation. The ``Avg1(Per)'' denotes the average percentage of knowledge preservation. The ``Avg2'' is the average performance of downstream tasks. ``GM'' is the geometric mean of Avg1 and Avg2.}
\label{llama3.2_ab_appendix}
\end{table*}
\section{More results}

Table \ref{tab:effecitve_ranks_complete} is the complete version of Table \ref{tab:effecitve_ranks}. Table \ref{llama3.2_ab_appendix} is the complete version of Table \ref{llama3.2_ab}. Table \ref{llama2_different_seeds_math} and \ref{llama2_different_seeds_code} add random seeds for \ref{llama2_math} and \ref{llama2_code}. Table \ref{llama2_different_seeds_math} and \ref{llama2_different_seeds_code} demonstrate that across varying random seeds, LoRA-Null consistently surpasses MiLoRA from CorDA in performance. Table \ref{llama2_more_code_result_2} shows the results for code task trained on another dataset, \text{i.e.},  WizardLM-Evol-Instruct\cite{pmlr-v235-wei24h}. Table \ref{llama2_more_code_result_2} further validates the effectiveness of our method. Table \ref{llama3_math_result} and Table \ref{llama3.1_math_result} present the results of the mathematical tasks on Llama-3-8B and Llama-3.1-8B, respectively. On Llama-3-8B, LoRA-Null achieves the best performance, while on Llama-3.1-8B, LoRA-Null is only slightly behind MiLoRA. Overall, LoRA-Null remains the best-performing method. These demonstrate that the space of LoRA initialization is the key to preserving pre-trained knowledge rather than the residual weights.

\begin{table*}[t!]
\centering
\normalsize % ✅ 显式设为 10pt Roman 字体

% --- Subtable (a): Code ---
\begin{subtable}{\linewidth}
\centering
\subcaption{Math}
\label{llama2_code_seeds}
\setlength{\tabcolsep}{3pt} % ✅ 减小列间距，替代 \resizebox
\begin{tabular}{l|c|ccccc|ccc|c}
    \toprule[1.5pt]
    Method & \textbf{\#Params} & Trivia QA & NQ open & WebQS & Avg1 & Avg1(per) & HumanEval & MBPP & Avg2 & GM  \\
    \midrule
    LLaMA-2-7B & - & 52.51 & 18.83 & 5.91 & 25.75  & 100.00 & - \\
    \midrule
    MiLoRA (seed 133)& 320M & 48.90 & 3.99 & 6.35 & 19.75 & 71.44 &  40.18 & 5.28 & 22.73 &  21.19\\
    
    CorDA (seed 133)& 320M &  49.46  & 3.27 & 6.20 & 19.64 & 70.52  & 38.29 & 5.04 & 21.67 & 20.63 \\
    
    LoRA-Null (seed 133)& 320M & \textbf{50.53} & 7.04 & \textbf{6.69} & \underline{21.42} & \underline{77.87} & 42.23 & 6.18 & 24.21 & 22.77 \\
    
    MiLoRA (seed 233) & 320M & 47.02 & 3.66 & 6.10 & 18.93 & 69.66 & 41.47 & 6.20 & 23.84 & 21.24 \\
    
    CorDA (seed 233) & 320M & 48.99 & \underline{7.15} & 5.76 & 20.63 & 76.24 & 41.47 & \underline{8.22} & 24.85 & 22.64 \\
    
    LoRA-Null (seed 233) & 320M & \underline{50.02} & \textbf{7.98} & \underline{6.55} & \textbf{21.52} & \textbf{79.21} & \textbf{44.43} & \textbf{8.80} & \textbf{26.62} & \textbf{23.93} \\
    
     MiLoRA (seed 333) & 320M & 48.92 & 4.10 & 6.45 & 19.82 &  71.65 & 39.88 & 4.84 & 22.73 & 21.22 \\
     
    CorDA (seed 333) & 320M & 49.55 & 5.10 & 6.15 & 20.27 & 73.82 & 37.91 & 5.00 & 21.46 & 20.86 \\
    
    LoRA-Null (seed 333) & 320M &  49.88 & 6.30 & 6.37 & 20.85 & 76.15 & \underline{43.29} & 6.10 & \underline{24.70} & \underline{22.69}  \\
    \bottomrule[1.5pt]
\end{tabular}
\end{subtable}

\caption{Performance evaluation of LoRA-Null on Math tasks compared with baseline methods. The first row is the original performance. \textbf{Bold} and \underline{underline} indicate top and runner-up results. The``Avg1'' is the average performance of preservation. The ``Avg1(Per)'' denotes the average percentage of preservation. The``Avg2'' is the average performance of downstream tasks.}
\label{llama2_different_seeds_math}
\end{table*}

\begin{table*}[t!]
\centering
\normalsize % ✅ 显式设为 10pt Roman 字体

% --- Subtable (a): Code ---
\begin{subtable}{\linewidth}
\centering
\subcaption{Code}
\label{llama2_math_seeds}
\setlength{\tabcolsep}{3pt} % ✅ 减小列间距，替代 \resizebox
\begin{tabular}{l|c|ccccc|ccc|c}
    \toprule[1.5pt]
    Method & \textbf{\#Params} & Trivia QA & NQ open & WebQS & Avg1 & Avg1(per) & HumanEval & MBPP & Avg2 & GM  \\
    \midrule
    LLaMA-2-7B & - & 52.51 & 18.83 & 5.91 & 25.75  & 100.00 & - \\
    \midrule
    MiLoRA (seed 133)& 320M &  48.99 & 11.11 & 7.14 & 22.41 & 84.10 & 16.97 & 20.40 & 18.69 & 20.47 \\
    CorDA (seed 133)& 320M &  48.91 & 11.08 & 6.64 & 22.21 & 84.00 & 17.03 & 20.21 & 18.62 & 20.33 \\
    LoRA-Null (seed 133)& 320M &  50.67 & \underline{13.21} & 6.64 & 23.51 & \textbf{88.88} & 18.04 & 22.76 & 20.40 & 21.90 \\
    MiLoRA (seed 233)& 320M & 48.92 & 9.53 & 7.78 & 22.08 & 81.26& 15.83 & 23.33 & 19.58 & 20.79 \\
    CorDA (seed 233) & 320M & \underline{50.94} & 11.33 & \textbf{8.91} & \textbf{23.73} & 85.73 & \underline{17.97} & \underline{23.88} & \underline{20.93} & \underline{22.29} \\
    LoRA-Null (seed 233)& 320M & \textbf{51.34} & 11.52 & \underline{8.17} & \underline{23.68} & \underline{86.32} & 17.13 & \textbf{25.21} & \textbf{21.17} &  \textbf{22.39} \\
    MiLoRA (seed 333) & 320M & 49.52 & 10.80 & 6.99 & 22.44 & 85.76 & 17.16 & 19.30 & 18.23 & 20.23 \\
     CorDA (seed 333) & 320M &  50.06 & 11.86 & 6.94 & 22.95 & 86.11 & 16.97 & 19.60 & 18.29 & 20.49 \\
    LoRA-Null (seed 333)& 320M & 50.58 & \textbf{13.24} & 6.94 & 23.59 & \textbf{88.88} & \textbf{17.80} & 22.76 & 20.40 & 21.94 \\
    \bottomrule[1.5pt]
\end{tabular}
\end{subtable}

\caption{Performance evaluation of LoRA-Null on Code tasks compared with baseline methods. The first row is the original performance. \textbf{Bold} and \underline{underline} indicate top and runner-up results. The``Avg1'' is the average performance of preservation. The ``Avg1(Per)'' denotes the average percentage of preservation. The``Avg2'' is the average performance of downstream tasks.}
\label{llama2_different_seeds_code}
\end{table*}
\begin{table*}[t!]
\centering
\normalsize % ✅ 显式设为 10pt Roman 字体

% --- Subtable (a): Code ---
\begin{subtable}{\linewidth}
\centering
\subcaption{Code}
\label{llama2_code_2}
\setlength{\tabcolsep}{3pt} % ✅ 减小列间距，替代 \resizebox
\begin{tabular}{l|c|ccccc|ccc|c}
    \toprule[1.5pt]
    Method & \textbf{\#Params} & Trivia QA & NQ open & WebQS & Avg1 & Avg1(per) & HumanEval & MBPP & Avg2 & GM  \\
    \midrule
    LLaMA-2-7B & - & 52.51 & 18.83 & 5.91 & 25.75  & 100.00 & - \\
    \midrule
    %Full fine-tuning & 6738M & 41.41 & 3.74 & 5.86 & 17.00 & 65.96 & 16.19 & 22.56 &  19.38 & 18.15 \\
    LoRA & 320M & 48.35 & 8.07 & 6.81 & 21.08 & 78.31 & 15.72 & 23.80 & 19.76&  20.41 \\
    PiSSA & 320M & 48.02 & 8.73 & 7.82 & 21.52 & 79.27  & \textbf{19.36} & \underline{24.81} & \textbf{22.09} &  21.80 \\
    MiLoRA & 320M & 48.92 & 9.53 & 7.78 & 22.08 & 81.26& 15.83 & 23.33 & 19.58 & 20.79 \\
    CorDA & 320M & \underline{50.94} & \underline{11.33} & \textbf{8.91} & \textbf{23.73} & \underline{85.73} & \underline{17.97} & 23.88 & 20.93 & \underline{22.29} \\
    LoRA-Null & 320M & \textbf{51.34} & \textbf{11.52} & \underline{8.17} & \underline{23.68} & \textbf{86.32} & 17.13 & \textbf{25.21} & \underline{21.17} &  \textbf{22.39} \\
    \bottomrule[1.5pt]
\end{tabular}
\end{subtable}

\caption{Performance evaluation of LoRA-Null on Code tasks using WizardLM-Evol-Instruct \cite{pmlr-v235-wei24h} compared with baseline methods. The first row is the original performance. \textbf{Bold} and \underline{underline} indicate top and runner-up results. The``Avg1'' is the average performance of preservation. The ``Avg1(Per)'' denotes the average percentage of preservation. The``Avg2'' is the average performance of downstream tasks.}
\label{llama2_more_code_result_2}
\end{table*}
\begin{table*}[t!]
\centering
\small % ✅ 显式设为 10pt Roman 字体

% --- Subtable (a): Math on LLaMA-3-8B ---
\begin{subtable}{\linewidth}
\centering
\subcaption{Math on LLaMA-3-8B}
\label{llama3_math}
\setlength{\tabcolsep}{1mm} % ✅ 减小列间距替代 \resizebox
\begin{tabular}{l|c|ccccc|ccc|c}
    \toprule[1.5pt]
    Method & \textbf{\#Params} & Trivia QA & NQ open & WebQS & Avg1& Avg1(Per)&  GSM8k & Math & Avg2 & GM \\
    \midrule
    LLaMA-3-8B & - & 63.33 & 13.77 & 13.73 & 30.28 & 100.00 & - & - & - & - \\
    \midrule
    %Full fine-tuning & 8,366M & 54.90 & 6.12 & 8.17 & 23.03 &  63.55 & \underline{76.88} & \underline{25.94} & \underline{51.41} & 34.41\\
    LoRA & 336M & \underline{60.67} & 8.34 & 12.30 & 27.10 & 81.98 & 72.86 & 24.32 & 48.59 & 36.29 \\
    PiSSA & 336M & 50.60 & 5.82 & 8.12 & 21.51 & 60.44 & \textbf{77.10} & \textbf{26.26} & \textbf{51.68} &  33.34 \\
    MiLoRA & 336M & 60.08 & \underline{9.11} & \textbf{12.99} & \underline{27.39} & \underline{85.21} & 73.24 & 23.90 & 48.57 &  \underline{36.47} \\
    CorDA & 336M & 59.68 & 7.89 & 10.53 & 26.03 & 76.08 & 74.98 & \underline{25.84} &  \underline{50.41} & 36.22 \\
    LoRA-Null & 336M & \textbf{62.00} & \textbf{9.97} & \underline{12.89} & \textbf{28.29} & \textbf{88.06} & \underline{75.06} & 24.92 &  49.99 & \textbf{37.61}\\
    \bottomrule[1.5pt]
\end{tabular}
\end{subtable}

\caption{Performance evaluation of LoRA-Null on Math tasks compared with baseline methods using LLaMA-3-8B. Experiments employ LLaMA-3-8B for fine-tuning. LoRA, PiSSA, MiLoRA, CorDA, and LoRA-Null-v1 utilize the same rank $r=128$. CorDA, LoRA-Null-v1, and LoRA-Null-v2 utilize the NQ open dataset to initialize their adapters. All methods were implemented using the same training and evaluation settings. The ``LLaMA-3-8B'' baseline row indicates the original model's world knowledge preservation. \textbf{Bold} indicates the best result; \underline{underlined} indicates the runner-up.}
\label{llama3_math_result}
\end{table*}
\begin{table*}[t!]
\centering
\small % ✅ 显式设为 10pt Roman 字体

% --- Subtable (a): Math on LLaMA-3.1-8B ---
\begin{subtable}{\linewidth}
\centering
\subcaption{Math on LLaMA-3.1-8B}
\label{llama3.1_math}
\setlength{\tabcolsep}{3pt} % ✅ 减小列间距替代 \resizebox
\begin{tabular}{l|c|ccccc|ccc|c}
    \toprule[1.5pt]
    Method & \textbf{\#Params} & Trivia QA & NQ open & WebQS & Avg1& Avg1(Per) & GSM8k & Math & Avg2 & GM \\
    \midrule
    LLaMA-3.1-8B & - & 61.54 & 7.75 & 11.86 & 27.05  & 100.00  & - & - & - & - \\
    \midrule
    %Full fine-tuning & 8,366M & 45.63 & 4.02 & 6.40 & 18.68 & 59.99 & \underline{77.94} & \underline{27.88} & 52.91 & 31.44\\
    LoRA & 336M & 55.95 & 4.49 & 9.84 & 23.43 & 77.27 & 73.46 & 26.26 &  49.86 & 24.18\\
    PiSSA & 336M & 46.88 & 5.01 & 7.14 & 19.68 & 67.01 & \textbf{78.09} &  \textbf{28.50} &  \textbf{53.30} & 32.39\\
    MiLoRA & 336M & \underline{57.31} & \textbf{6.18} & \textbf{11.22} & \textbf{24.90} & \textbf{89.16} & 73.46 & 26.28 & 49.87 & \textbf{35.24} \\
    CorDA & 336M & 56.35 & \underline{5.37} & 8.56 & 23.42& 77.68 & 76.35 & \underline{27.78} &  \underline{52.07} & 34.92\\
    LoRA-Null & 336M & \textbf{57.38} & 4.29 & \underline{10.24} & \underline{23.97} & \underline{78.31} & \underline{75.06} & 27.68 &  51.37 & \underline{35.09}\\
    \bottomrule[1.5pt]
\end{tabular}
\end{subtable}

\caption{Performance evaluation of LoRA-Null on Math tasks compared with baseline methods using LLaMA-3.1-8B. The first row is the original performance. \textbf{Bold} and \underline{underline} indicate top and runner-up results. The``Avg1'' is the average performance of preservation. The ``Avg1(Per)'' denotes the average percentage of preservation. The``Avg2'' is the average performance of downstream tasks. }
\label{llama3.1_math_result}
\end{table*}

\end{document}